%% file: main.tex
\documentclass{article} 
\usepackage{iclr2026_conference,times}

\input{math_commands.tex}

\usepackage{booktabs}        
\usepackage{multirow}
\usepackage{makecell}        
\usepackage{threeparttable}  
\usepackage{siunitx}         
\usepackage{graphicx}   
\usepackage[table]{xcolor} 
\usepackage{amsmath}      
\usepackage{fontawesome5} 
\usepackage{subcaption}
\usepackage{pifont}
\usepackage{wrapfig}
\usepackage{caption}

\definecolor{BlueWash}{HTML}{EAF2FF}   
\definecolor{CyanWash}{HTML}{E6FAF7}   
\definecolor{MintWash}{HTML}{EDF7EE}   
\definecolor{AmberWash}{HTML}{FFF4DB}  
\definecolor{RoseWash}{HTML}{FDECF1}   
\definecolor{LavenderWash}{HTML}{F2ECFF}
\definecolor{SlateWash}{HTML}{F2F4F8}  

\definecolor{lightgray}{gray}{0.9}
\definecolor{myred}{RGB}{214, 39, 40} 
\newcommand{\imprv}[1]{\textcolor{myred}{\scriptsize($\uparrow$#1)}}

\usepackage{listings}
\usepackage{xcolor}

\lstdefinelanguage{json}{
    basicstyle=\ttfamily\footnotesize,
    numbers=left,
    numberstyle=\tiny\color{gray},
    stepnumber=1,
    numbersep=5pt,
    showstringspaces=false,
    breaklines=true,
    frame=single,
    backgroundcolor=\color{gray!5},
    literate=
     *{0}{{{\color{blue}0}}}{1}
      {1}{{{\color{blue}1}}}{1}
      {2}{{{\color{blue}2}}}{1}
      {3}{{{\color{blue}3}}}{1}
      {4}{{{\color{blue}4}}}{1}
      {5}{{{\color{blue}5}}}{1}
      {6}{{{\color{blue}6}}}{1}
      {7}{{{\color{blue}7}}}{1}
      {8}{{{\color{blue}8}}}{1}
      {9}{{{\color{blue}9}}}{1}
      {:}{{{\color{red}:}}}{1}
      {,}{{{\color{red},}}}{1}
      {"}{{{\color{black}"}}}{1},
}

\usepackage{float}  
\newcommand{\wPone}{w_{\text{P1}}}   
\newcommand{\wPtwo}{w_{\text{P2}}}   
\newcommand{\apred}{a_{\mathrm{pred}}}   
\newcommand{\rmin}{r_{\mathrm{min}}}     

\usepackage{hyperref}
\usepackage{url}
\usepackage[ruled,vlined,linesnumbered]{algorithm2e}
\usepackage{dsfont}

\usepackage{amssymb}

\title{K-frames: Scene-Driven Any-k Keyframe Selection for long video understanding}

\author{
Yifeng Yao\textsuperscript{\normalfont 1,2}, 
Yike Yun\textsuperscript{\normalfont 2}, 
Jing Wang\textsuperscript{\normalfont 1}, 
Huishuai Zhang\textsuperscript{\normalfont 1}, 
Dongyan Zhao\textsuperscript{\normalfont 1,}\thanks{Corresponding Author zhaodongyan@pku.edu.cn},\\
\textbf{
Ke Tian\textsuperscript{\normalfont 2}, 
Zhihao Wang\textsuperscript{\normalfont 2}, 
Minghui Qiu\textsuperscript{\normalfont 2}, 
Tao Wang\textsuperscript{\normalfont 2}}\\
\{yaoyifeng, wangjing\}@stu.pku.edu.cn, yike.yun@bytedance.com\\
\textsuperscript{\normalfont 1}Wangxuan Institute of Computer Technology, Peking University, \textsuperscript{\normalfont 2}Bytedance
}

\iclrfinalcopy 
\begin{document}

\maketitle
\begin{abstract}

Multimodal Large Language Models (MLLMs) have demonstrated significant capabilities in image understanding, but long-video are constrained by context windows and computational cost. Uniform frame sampling often leads to substantial information loss. Meanwhile existing keyframe selection methods such as text-frame retrieval or RL-based frame optimization typically yield sparse and temporally disjointed frames, overlooking scene continuity and lacking flexibility for multi-scale frame selection. To address these limitations, we introduce K-frames, a novel paradigm for scene-driven keyframe selection that preserves temporal continuity. Instead of selecting individual frames, K-frames predicts semantically coherent, query-relevant clips, which enables any-k keyframes selection to meet diverse user budgets. To achieve this approach, we first introduce PeakClips, a dataset of 200K video highlights conditioned by query. Building on this dataset, K-frames learns clip2frame selection using a three-stage progressive curriculum. It involves two Supervised Fine-Tuning stages for temporal grounding and key-clip perception, followed by a Reinforcement Learning stage that directly optimizes the scene-driven prediction policy for downstream task without further annotations. Extensive experiments on major long-video understanding benchmarks demonstrate that K-frames provides an effective, interpretable, and plug-and-play solution for keyframe selection at various scales. Our dataset and model will be available.

\end{abstract}

\section{Introduction}
Recent progress in Multimodal Large Language Models  (MLLMs)~\citep{qwen2.5vl, wang2025internvl3} has come from coupling Large Language Models (LLMs) with vision encoders via a cross-modal projector that maps visual features into the language token space.  This design enables unified, instruction-following multimodal reasoning across diverse text-image tasks. However, extending these models from image to video remains challenging. As treating a video as a sequence of frames greatly increases the number of visual tokens, especially for long videos. On the one hand, finite context windows cannot accommodate all video frames. On the other hand, the quadratic computational complexity of standard Transformer attention~\citep{vaswani2017attention} makes longer inputs dramatically more expensive in computation and in token-metered API usage. Therefore, frame downsampling is practically necessary for video inputs.

Current MLLMs typically process videos via uniform frame sampling. But for long videos, the challenge is that sampling only a small subset of frames risks a critical loss of context, highlighting the need for keyframe selection. Existing methodologies for keyframe selection are predominantly categorized into two paradigms: text-frame retrieval and Reinforcement Learning (RL)-based optimization. The former computes the similarity of frames and text query to rank frames~\citep{AKS}, treating video as independent images. This neglects temporal context and struggles with instruction-heavy or compositional queries. The latter, RL-based methods, optimize frame subsets for downstream objectives. But the resulting selections are typically sparse, which harms scene continuity, thereby degrading video understanding performance. And it also fails to accommodate personalized user budgets due to the lack of flexibility for multi-scale selection.

\begin{figure*}[!t]
\centering
\includegraphics[width=1\textwidth]{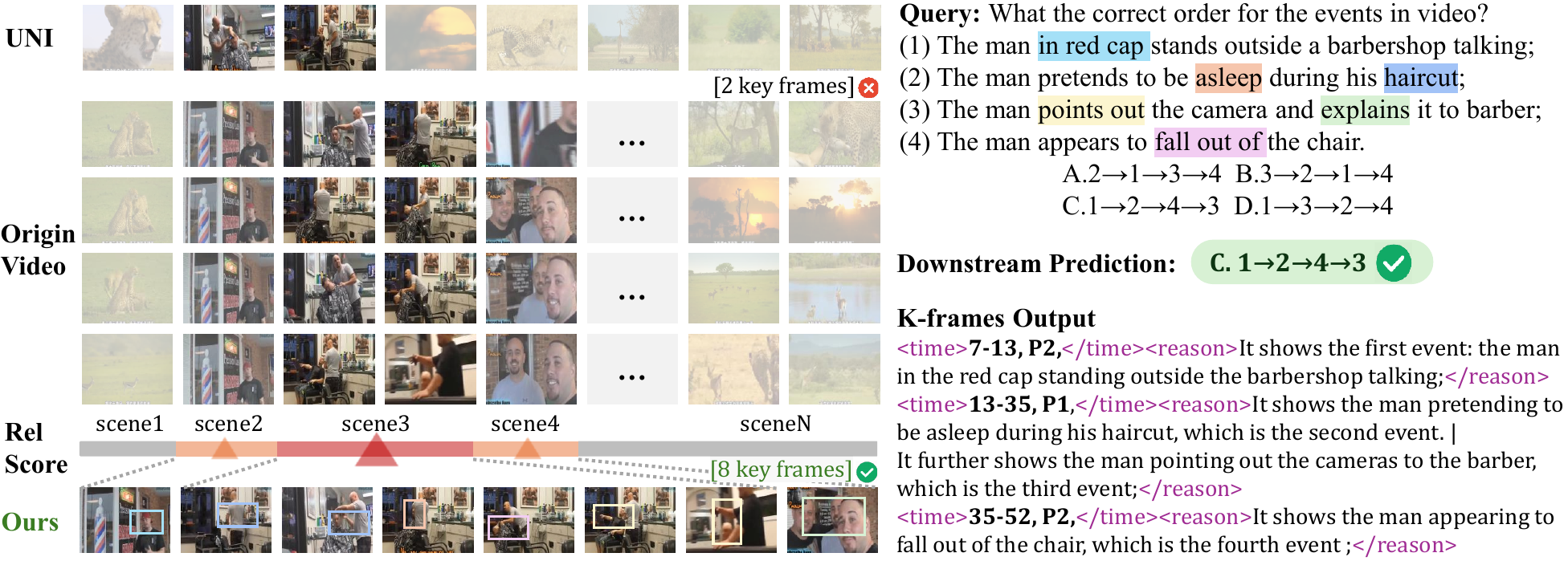} 
\caption{Visualization of our K-frames paradigm. Unlike uniform sampling (UNI), our model first predicts query-relevant key clips along the video timeline, assigning them importance levels of P1 (top-priority) or P2 (secondary-priority). Keyframes are then selected based on these key clips.}
\label{fig:intro}
\vspace{-2ex}
\end{figure*}

To address these limitations, we propose K-frames, a query-conditioned and interpretable paradigm that reframes keyframe selection as clip2frame prediction. Instead of selecting isolated frames, K-frames first localizes semantically coherent, temporally contiguous clips aligned with the query, and then selects any-k keyframes based on those clips. As illustrated in Figure~\ref{fig:intro}, this clip-first design preserves scene continuity, focuses computation on informative regions, making the selection process interpretable. As a model-agnostic front-end, K-frames enhances the efficiency and performance of existing MLLMs in long video understanding with no modifications to their architecture.

The main challenge in scene-driven keyframe selection is the lack of scene-level relevance annotations. To close this gap, we construct a new dataset, \textbf{PeakClips}, with hierarchical captions and detailed video highlight annotations. PeakClips is built via a three-stage pipeline: (1) scene segmentation partitions videos into scene-aware temporal units based on changes in visual content.; (2) hierarchical captioning at the scene/chapter/video levels supplies multi-granular descriptions that link local scene to the global narrative; and (3) LLM-guided relevance scoring aligns scenes with the query through Gemini 2.5 Pro~\citep{comanici2025gemini}, and using frame–query similarity further refines relevance score to the frame level. By annotating these scenes, we ultimately aim to supply keyframe selection, temporal localization, and hierarchical understanding in long-term video.

Building on the PeakClips dataset, we employ a three-stage progressive  curriculum to tecach K-frames. We use a lightweight MLLM (Qwen2.5-VL-3B) as the backbone. The initial Supervised Fine-Tuning (SFT) stage prepares the model for our scene-driven paradigm by instilling foundational capabilities in temporal localization and scene understanding. Then during the second SFT stage the model learns with supervised data to perceive query-relevant video clips with reason, enabling our clip2frame prediction. Finally, the SFT-trained model serves as a cold-start policy for Reinforcement Learning, where the scene-driven keyframe selection policy is directly optimized to ensure the selected scenes are maximally effective for downstream task. This entire process yields a model that outputs query-conditioned key clips rather than disconnected frames, naturally enabling interpretable and flexible any-k keyframe selection.


To sum up, the main contributions are: 
(1) We construct PeakClips, a 200K query-conditioned highlight dataset built via scene segmentation, hierarchical captioning, and LLM-guided relevance scoring, providing supervision for temporal grounding, scene perception, and keyframe prediction. 
(2) We propose K-frames, a new interpretable paradigm that reframes keyframe selection as clip2frame prediction, preserving scene continuity and enabling any-k keyframe selection.
(3) Extensive experiments on major long-video understanding benchmarks demonstrate that K-frames provides an effective, interpretable, and plug-and-play solution for keyframe selection at multi-scales. 

\section{Related Work}
\subsection{Multi-modal Large Language Models for Video Understanding}
Existing MLLMs such as ChatGPT-4o~\cite{gpt4o}, Gemini 2.5 Pro~\cite{comanici2025gemini} and Qwen-VL 2.5~\cite{qwen2.5vl} have made significant progress in multimodal understanding~\citep{gpt4,gemini,qwen-vl}. However, adapting these models to the video domain introduces the added complexity of modeling temporal information. Early efforts in video-MLLMs primarily relied on uniformly sampled frames and simple connectors, such as MLPs~\citep{video-llava,minigpt4-video,video-chatgpt}, discrete visual tokenizers~\citep{video-lavit} and Q-formers~\citep{video-llama,mvbench} to link visual encoders with LLMs. Subsequent models focus on enhanced video-instruction data~\citep{llava-next,internvideo2}, efficient spatio-temporal feature compression methods~\citep{longvu,koala} and video-specific encoders~\citep{internvideo2,videogpt+}. 
Specifically, processing long videos remains a significant bottleneck due to MLLMs' context limits and prohibitive computational costs. Current strategies to mitigate this challenge include directly extending the LLM's context window~\citep{longva}, developing memory management mechanisms~\citep{malmm} or keyframe selection algorithms~\citep{AKS,refocus,viarl} for identifying representative frames. 

\subsection{Existing Keyframe Selection Methods}
Efficient keyframe selection has become a critical component for long-video understanding, evolving from traditional approaches like query-agnostic clustering-based methods~\citep{zhang2013efficient} or uniform sampling~\citep{pllava} to modern query-adaptive strategies. They are predominantly divided into two paradigms: text-image retrieval and RL-based frame optimization. Text-image retrieval methods calculate the independent video frame-query similarity to localize important frames. 
MLLM Based Frame Selection~\citep{MLLM-selection} employs spatial-temporal importance scoring to boost performance, and 
Frame-Voyager~\citep{voyager} ranks frame combinations via pretrained Video-LLMs. 
Concurrently, there have been endeavors to integrated RL into keyframe selection for policy optimization. ReFoCUS~\citep{refocus} proposed a frame-level policy optimization framework that shifts the optimization target from textual responses to visual input selection, and ViaRL~\citep{viarl} leverages the downstream model's answer accuracy as a reward signal, enabling a trial-and-error learning that requires no explicit frame selection annotations. 
Yet, these approaches prioritize frame-level semantics, largely ignoring a video's crucial temporal structure.
In contrast, our method K-frames redefines this task through clip2frame prediction, a paradigm that preserves the narrative flow of events and supports versatile any-k selection.

\section{Method}

\begin{figure*}[!t]
\centering
\includegraphics[width=1\textwidth]{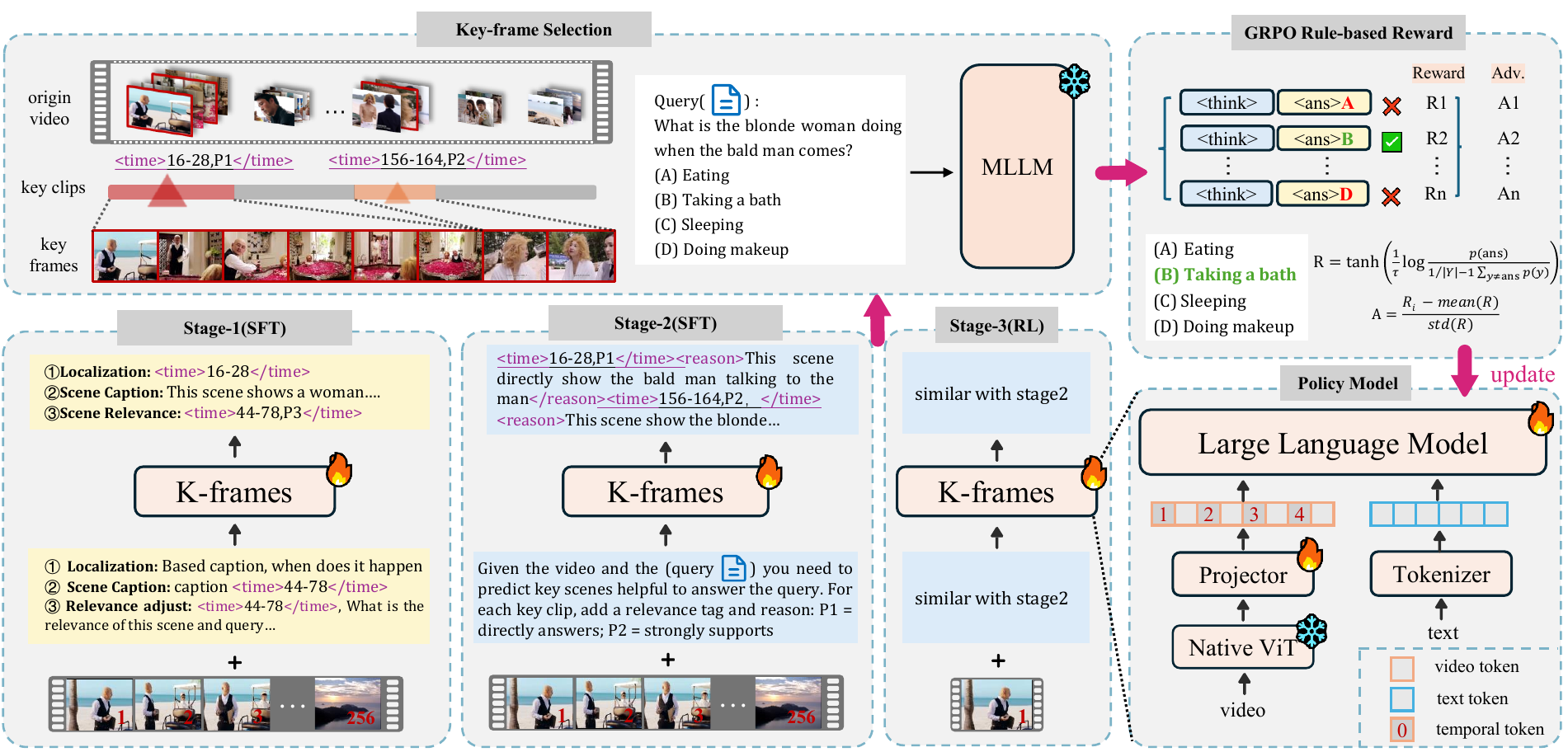} 
\caption{An overview of the K-frames framework. It features a two-stage Supervised Fine-Tuning (SFT) curriculum for temporal grounding and key-clip perception, followed by a Reinforcement Learning (RL) stage to align the selection policy with downstream task performance.}
\label{fig:main_figure}
\vspace{-1ex}
\end{figure*}

In this work, we propose K-frames, which reframes keyframe selection as the task of predicting query-relevant key clips and sampling frames. To achieve this, our model needs to understand scene-level semantics and their temporal boundaries. A main challenge, however, is the lack of datasets with scene-level relevance annotations. To address this, we first present the construction of our large-scale dataset, PeakClips, which provides the necessary supervision (Sec. \ref{sec:sup}). Building on this dataset, we train K-frames using a novel three-stage progressive curriculum. We begin with two stages of Supervised Fine-Tuning to equip the model with the fundamental capabilities of temporal grounding and key-clip perception (Sec. \ref{sec:key_clip}). Finally, we employ Reinforcement Learning to align the model's clip2frame selection policy with downstream long-video understanding tasks, without the need for further annotations (Sec. \ref{sec:anno}). The overall system is illustrated in Figure \ref{fig:main_figure}.

\begin{figure*}[!t]
\centering
\includegraphics[width=1\textwidth]{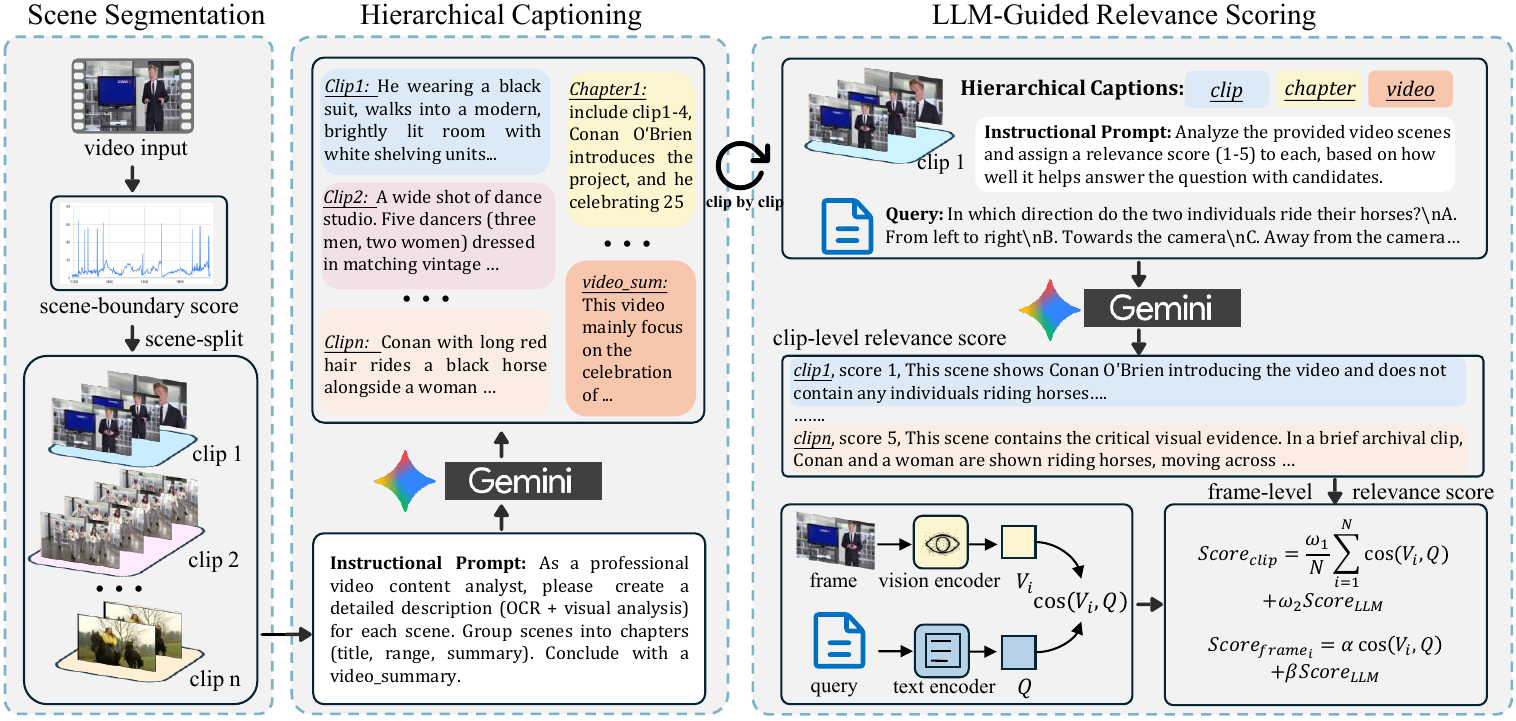} 
\caption{The three-stage framework for constructing the PeakClips dataset. The process involves (1) Scene-aware Segmentation to partition the video, (2) Hierarchical Captioning to generate multi-level descriptions, and (3) LLM-guided Relevance Scoring to identify query-conditioned relevance.}
\label{fig:dataset_construction}
\vspace{-1ex}
\end{figure*}

\subsection{PeakClips Dataset}
\label{sec:sup}
To enable our clip-to-frame learning paradigm, we introduce \textbf{PeakClips}, a large-scale dataset comprising over 200K query-conditioned relevance annotations on video clips. The videos used for annotation are sourced from LLaVA-Video-178K~\citep{zhang2024video}, NeXT-QA~\citep{xiao2021next}, PerceptionTest~\cite{patraucean2023perception} datasets. As illustrated in Figure~\ref{fig:dataset_construction}, the construction follows a three-stage pipeline to assess the relevance of all clips within a video $v$ to a given query, ultimately generating a set of key clips $\{clip_v^i = [frame\_start_v^i, frame\_end_v^i]\}$.

\paragraph{Scene-aware Segmentation.}
We first decompose each video into a set of temporally contiguous and semantically coherent scenes. To achieve this, we calculate the change in visual content throughout the video by computing the histogram difference between consecutive frames~\citep{sheena2015key}. This process generates a scene-boundary score for each frame transition, where high scores correspond to abrupt changes in visual content. By segmenting the video at these high-scoring boundaries $\{b_v^0=1, b_v^1,\dots,b_v^M\}$, we obtain a set of scene clips $s_v^j = {[b_v^{j-1}, b_v^j]}$.

\paragraph{Hierarchical Captioning.}
To provide multiscale context for relevance scoring, we generate captions through Gemini 2.5 Pro~\citep{comanici2025gemini} at three granularities: \textbf{fine-grained clip-level descriptions}, \textbf{chapter-level summaries} (grouping related clips), and a \textbf{video-level synopsis}. This hierarchy allows relevance to be assessed by connecting local events to the global narrative, which is crucial for handling complex queries.

\paragraph{LLM-guided Relevance Scoring.}
With the segmented clips and captions, we first use Gemini 2.5 Pro to assign a base relevance score (1-5) with a reason to each clip based on a detailed instructional prompt (see Appendix~\ref{appendix:dataset}). This LLM-generated score is then refined using the text-frame similarity. Specifically, we compute a final clip-level score by taking a weighted average of the LLM score and the mean SIGLIP similarity between the query and the clip's frames. We are able to extend the relevance score to frame level by weighting the parent clip's Gemini score with each frame's individual SIGLIP-query similarity. But in our work, we only use the clip-level relevance. Clips with a final score greater than or equal to 4.9 are annotated as top-priority (\textbf{P1}) highlights, while those with scores in the range $[4.3, 4.9)$ are labeled as secondary-priority (\textbf{P2}) clips.


\paragraph{PeakClips Dataset.}
In summary, the three-stage construction pipeline yields the \textbf{PeakClips} dataset, a comprehensive resource for video understanding. Each entry provides videos annotated with temporally coherent scene boundaries, multi-level hierarchical captions (clip, chapter, and video), and query-conditioned highlight clips. The dataset also includes the dense, continuous clip-level relevance scores and LLM-generated rationales that informed the final selections. Collectively, these rich annotations make PeakClips a versatile resource for supervising a wide spectrum of tasks, including temporal grounding, scene-level perception, keyframe selection.

\subsection{Supervised fine-tuning for key-clip prediction}
\label{sec:key_clip}
Building on the PeakClips dataset, K-frames learns scene-driven keyframe selection through a three-stage progressive curriculum~\citep{bengio2009curriculum}. As illustrated in Figure~\ref{fig:main_figure}, it includes two-stage Supervised Fine-Tuning (SFT) and one-stage Reinforcement Learning (RL). The first two stages is to equip a lightweight MLLM(Qwen2.5-VL-3B) with two core capabilities essential for our task: temporal localization, and query-conditioned key clips perception. 

\paragraph{Temporal Grounding and Relevance Judge.}
In the first SFT stage, K-frames leverage the hierarchical captions and clip-query relevance annotations in PeakClips to learn Temporal Grounding. To enhance the K-frames' ability to align visual content with its time span, we employ two temporal prompting techniques throughout all three stages training. Following prior work \citep{wu2025number}, Visual Prompts render the frame index $t$ directly onto each frame $f_t$, providing a direct visual cue for time. Concurrently, we inject Textual Prompts preceding the visual tokens $\mathbf{v}_{t, t+1}$ for each frame. Building on these temporal cues, our curriculum is designed to instill robust localization and perception abilities. 

To directly enhance the model's temporal localization capabilities, we design a \textbf{caption-to-scene localization} task, where the model receives the video and a scene description to locate its temporal span. As a dual task, we introduce a \textbf{scene-to-caption generation}, requiring the model to generate a description for a given temporal span. Moreover, we incorporate a \textbf{clip-query relevance scoring} task. In this task, the model is required to predict how relevant a specific clip from the whole video is to a given query. The full specific instruction prompt for these three tasks see Appendix~\ref{sec:appendix_prompt}.

\paragraph{Query-Conditioned Key-Clip Prediction.} 

Building upon the foundational abilities learned in stage 1, the second SFT stage teach our model with parameters $\theta$ for its ultimate goal: given a long video $V = \{f_t\}_{t=1}^T$ and a query $Q$, it learns to perceive and predict a set of relevant key clips $\mathcal{C} = \{c_i\}_{i=1}^N$. Each predicted clip $c_i$ consists of a temporal span and a textual rationale. In this training phase, the model is conditioned on the full video $V$, the query $Q$, and a specific instruction prompt $I$. The prompt instructs the model to select query-relevant video clips, assigning a priority tag (\textbf{P1} for direct answers, \textbf{P2} for strong support), and providing a brief rationale for each selection (see Appendix~\ref{sec:appendix_prompt} for the full prompt text).

The training in both SFT stages is unified under a standard auto-regressive language modeling objective. The model is optimized to maximize the likelihood of generating the ground-truth sequence $\mathcal{Y}_{\text{gt}}$~\cite{mao2023cross}:
\begin{equation}
\mathcal{L}_{\text{SFT}} = -\log P(\mathcal{Y}_{\text{gt}} | V, Q, I; \theta)
\label{eq:sftloss}
\end{equation}
This holistic training compels the model to predict key clips conditioned on query. This process yields a well-initialized policy for the subsequent Reinforcement Learning stage and provides a strong, standalone model for key clip selection.

\subsection{Reinforcement Learning for Downstream Task Alignment}
\label{sec:anno}
To bridge the gap between mimicking annotations from Supervised Fine-Tuning and maximizing downstream task performance, we introduce a Reinforcement Learning stage. This stage directly optimizes the K-frames policy by aligning it with the final task objective, using the SFT-trained model as the initial policy. This alignment process requires no further annotations.

\paragraph{Scene-driven Keyframe Selection.}
The RL process begins with our SFT-trained K-frames, which functions as the actor model. For a given video $V$ and query $Q$, the actor model predicts a set of key clips. From these predicted clips, which represent the most informative segments, we then sample a fixed budget of $k$ keyframes using uniform sampling. This clip-first, sample-second strategy ensures that the selected frames are both semantically relevant and temporally coherent. These $k$ keyframes, along with the original query, are then fed into a powerful, frozen downstream MLLM (Qwen2.5-VL-7B) to generate a final answer to the query. The goal of our RL curriculum is to optimize the actor's clip2frame selection policy to maximize the quality of this final answer.

\paragraph{Policy Optimization with GRPO.}
To optimize our scene-driven keyframe selection policy, we employ Group Relative Policy Optimization (GRPO) \citep{grpo}, which eliminates the need for an explicit critic model by rolling out multiple candidate key clip selections and estimating their relative advantages. Instead of relying on a separate reward model, we compute a reward signal directly from the downstream model's output using a rule-based reward function. We perform this RL optimization exclusively on multiple-choice question-answering datasets to ensure a stable and reliable reward signal. The reward function evaluates answer quality by comparing the log-probability of the correct token against the average log-probability of incorrect ones, smoothed via a $\tanh$ transformation and a temperature hyperparameter $\tau$:
\begin{equation}
\mathrm{Reward} = \tanh\!\left( \frac{1}{\tau} 
\log \frac{p(\text{ans})}{\tfrac{1}{|Y|-1}\sum_{\hat{y} \neq \text{ans}} p(\hat{y})} \right)
\label{eq:reward}
\end{equation}
where $|Y|$ is the size of the candidate answer set and the probabilities $p(\cdot)$ are from the frozen downstream MLLM. To further improve training stability, we adopt the Dr. GRPO \citep{dr_grpo} variant, which modifies the advantage calculation and the overall optimization targets.

\begin{figure}[!b]
    \centering
    
    \begin{subfigure}[b]{0.32\textwidth}
        \includegraphics[width=\linewidth]{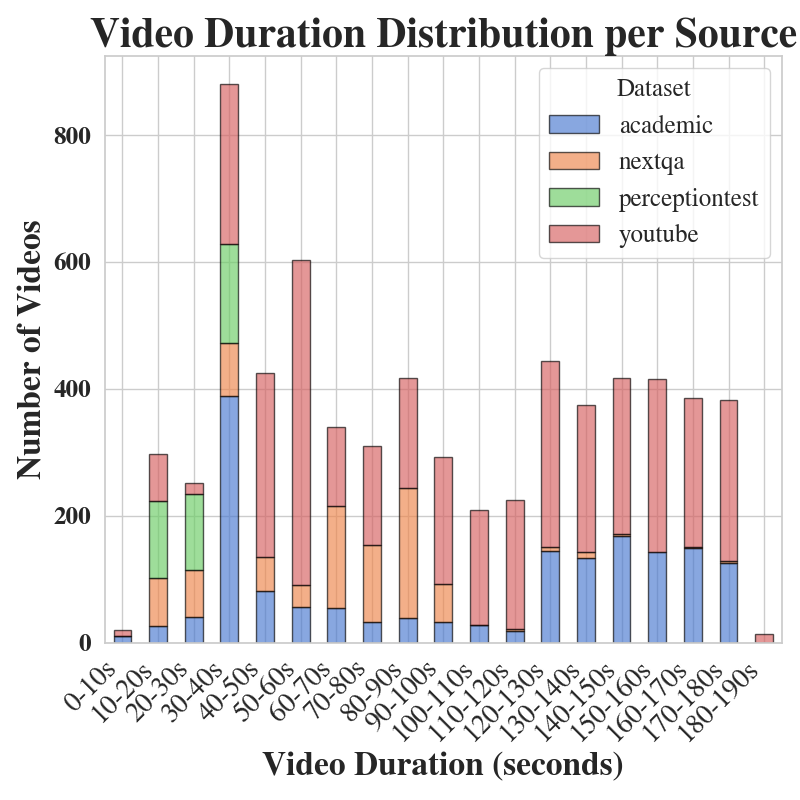}
        \caption{Video duration distribution.}
    \label{fig:duration}
    \end{subfigure}
    \hfill
    \begin{subfigure}[b]{0.32\textwidth}
        \includegraphics[width=\linewidth]{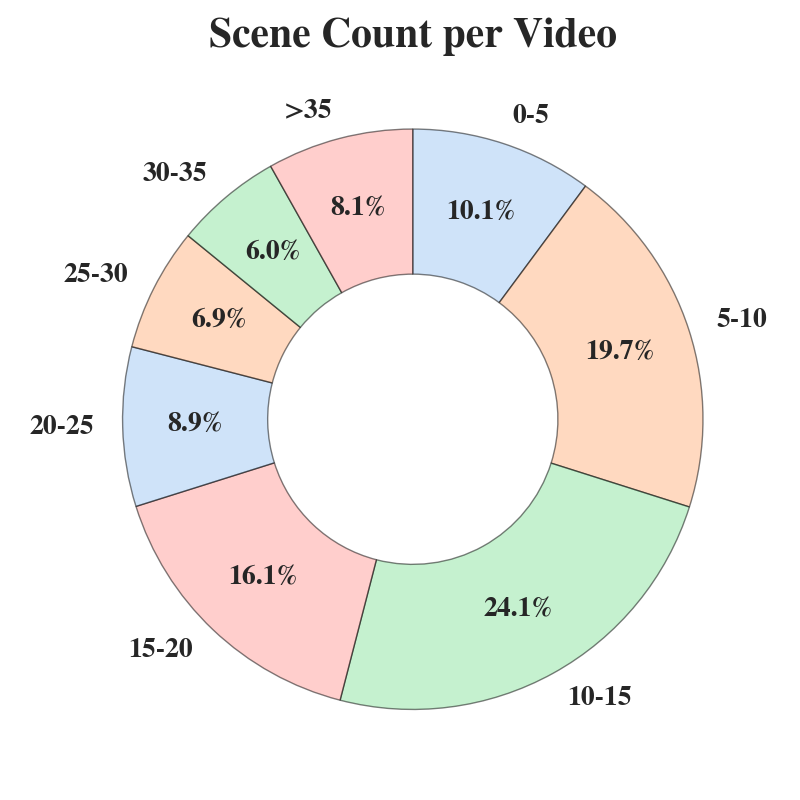}
        \caption{Scene count distribution.}
    \label{fig:count}
    \end{subfigure}
    \hfill
    \begin{subfigure}[b]{0.32\textwidth}
        \includegraphics[width=\linewidth]{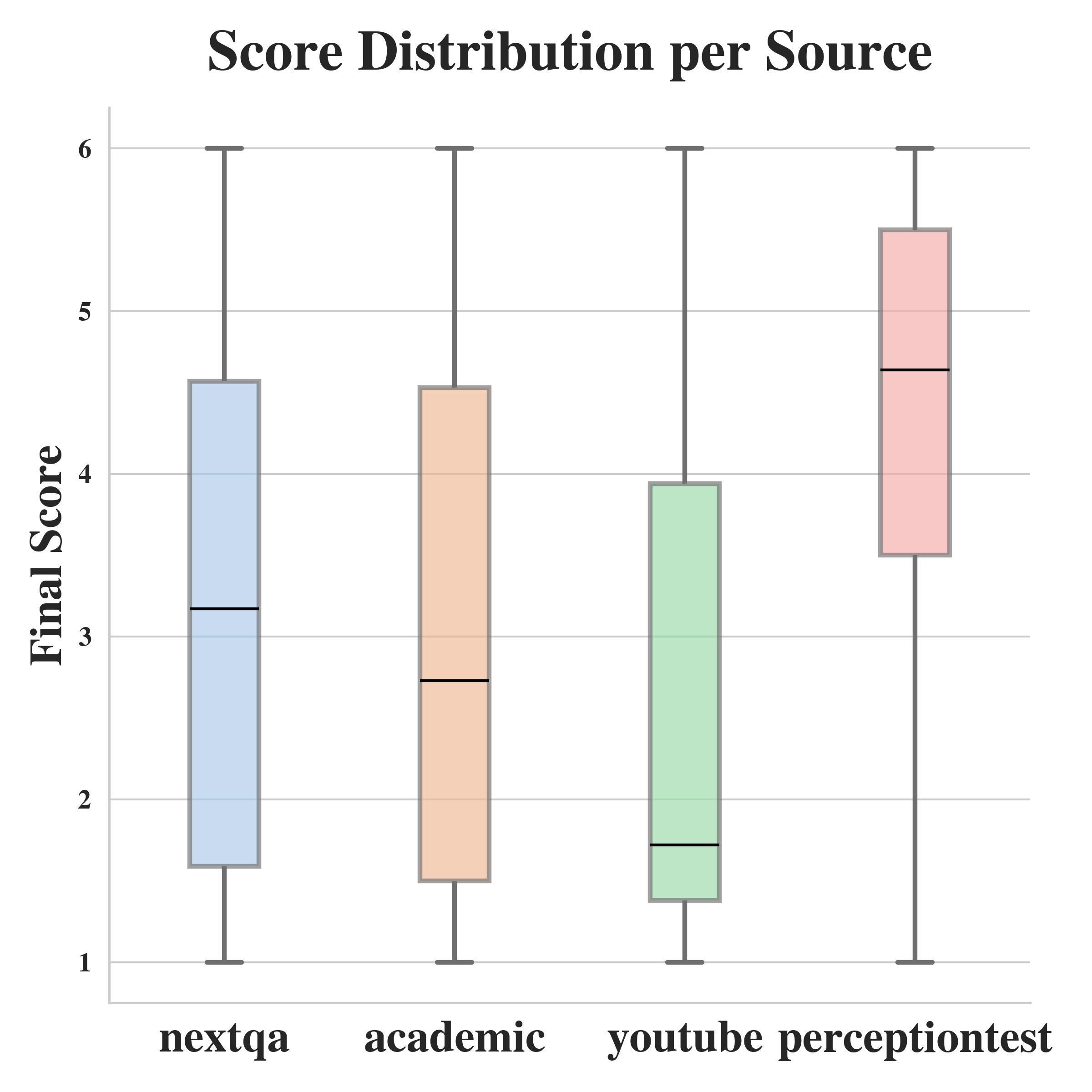}
        \caption{Scene score distribution.}
    \label{fig:score}
    \end{subfigure}
    \caption{Statistics about our proposed PeakClips dataset.}
    \vspace{-2ex}
\label{dataset}
\end{figure}

\paragraph{Any-K Keyframe Sampling.}
The full training pipeline yields the final, optimized K-frames model, which serves as a versatile, plug-and-play model for long-video understanding at inference. The process begins with K-frames predicting query-relevant key clips. Based on these predictions, we support two flexible keyframe sampling strategies: Focused Sampling (exclusively from key clips) and Hybrid Sampling (densely from key clips, sparsely from the rest of the video). This provides a flexible trade-off between deep focus and broad context. More detailes see in Appendix \ref{appd:infer}. These two strategies offer a flexible trade-off between concentrating on critical moments and maintaining broader video context.

\section{Experiments}
\subsection{PeakClips Statistics}

The \textbf{PeakClips} dataset consists of more than 200k annotations, derived from 6,702 randomly selected videos from the LLAVA-Video-178K \citep{llava-video} and labeled by Gemini 2.5 Pro. Specifically, our dataset consists of 108,221 scenes with 281,643 corresponding scene-query relevance annotations, and 19,070 chapters, with an average of 16.15 scenes and 2.85 chapters per video. As shown in Figure~\ref{fig:duration}, video durations vary notably across sources: PerceptionTest and NextQA clips are generally short, typically below 100 seconds, while Academic and YouTube videos exhibit a wider range with both short and long instances represented. Figure~\ref{fig:count} illustrates the scene count distribution per video: the majority fall within 5--15 scenes (43.8\%), while 10.1\% contain fewer than 5 scenes and 8.1\% exceed 35 scenes, indicating a balanced decomposition into semantically meaningful units. Finally, Figure~\ref{fig:score} reports the scene--query relevance scores produced by the LLM. PerceptionTest videos achieve the highest consistency with a median above 4.5, YouTube clips show broader variance with lower medians, and Academic and NextQA datasets lie between these extremes. For more dataset statistics please refer to Appendix \ref{appendix:dataset}.

\begin{table*}[t]
\centering
\small
\setlength{\tabcolsep}{3.5pt}
\setlength{\aboverulesep}{0pt}
\arrayrulecolor{black}
\caption{Main results on long-video understanding benchmarks. Our method (K-frames) consistently improves the performance of various open-source and closed-source MLLMs across different frames. The red text indicates the performance improvement over the baseline (uniform smpling). And the purple background highlights the largest improvement over the baseline.}
\begin{tabular}{@{\hspace{\tabcolsep}} c c c c c c c c c c @{\hspace{\tabcolsep}}}
\toprule
\multirow{2}{*}{Models} & \multirow{2}{*}{Size}& {\multirow{2}{*}{Frames}} & \multicolumn{2}{c}{MLVU} & \multicolumn{4}{c}{VideoMME} & {\multirow{2}{*}{LVBench}}\\
\cmidrule(lr){4-5}
\cmidrule(lr){6-9}
& & & {Needle-QA} & {M-Avg} & {Short} & {Medium} & {Long} & {Avg} & \\
\midrule
\rowcolor{lightgray}
\multicolumn{10}{c}{open-sourced model} \\
\midrule
Qwen2.5-VL & 7B & 8 & 58.6 & 53.9 & 61.7 & 50.6 & 46.9 & 53.0 & 52.8 \\
\textbf{+ ours} & 7B & 8 & \multicolumn{1}{c}{ 77.5 \bfseries\imprv{18.9}} & \multicolumn{1}{c}{ 60.4 \bfseries\imprv{6.5}} &68.9  & 55.3 & 47.9 & \multicolumn{1}{c}{\cellcolor{LavenderWash} 57.4 \bfseries\imprv{4.4}} & \multicolumn{1}{c}{ 57.7 \bfseries\imprv{4.9}} \\
\midrule
Qwen2.5-VL & 7B & 32 & 63.4 & 61.7 & 71.8 & 60.8 & 50.1 & 60.2 & 59.3 \\
\textbf{+ ours} & 7B & 32 &\multicolumn{1}{c}{79.4 \bfseries\imprv{16.0}} & \multicolumn{1}{c}{65.9 \bfseries\imprv{4.2}} & 74.1  & 61.4  & 51.7  & \multicolumn{1}{c}{62.1 \bfseries\imprv{1.9}} & \multicolumn{1}{c}{ 60.5 \bfseries\imprv{1.2}} \\
\midrule
Qwen2.5-VL & 7B & 64 & 67.7 & 65.6 & 73.9 & 62.3 & 52.2 & 62.8 & 59.9 \\
\textbf{+ ours} & 7B & 64 &\multicolumn{1}{c}{78.9 \bfseries\imprv{11.2}} & \multicolumn{1}{c}{ 67.8 \bfseries\imprv{2.2}} & 75.9 & 63.7 & 53.9 & \multicolumn{1}{c}{ 64.5 \bfseries\imprv{1.7}} & \multicolumn{1}{c}{ 61.6 \bfseries\imprv{1.7}} \\
\midrule
Qwen2.5-VL & 72B & 8 & 51.6 & 56.3 & 65.6 & 56.6 & 51.1 & 57.7 & 55.6 \\
\textbf{+ ours} & 72B & 8 &\multicolumn{1}{c}{77.2 \bfseries\imprv{25.6}} & \multicolumn{1}{c}{\cellcolor{LavenderWash} 63.3 \bfseries\imprv{7.0}} & 70.2 & 58.9  & 52.8 & \multicolumn{1}{c}{ 60.6 \bfseries\imprv{2.9}} & \multicolumn{1}{c}{ 59.3 \bfseries\imprv{3.7}} \\
\midrule
Qwen2.5-VL & 72B & 32 & 67.3 & 64.0 & 74.3 & 63.4 & 58.1 & 65.3 & 60.8 \\
\textbf{+ ours} & 72B & 32 &\multicolumn{1}{c}{ 78.3 \bfseries\imprv{11.0}} & \multicolumn{1}{c}{ 67.6 \bfseries\imprv{3.6}} & 75.2 & 66.0 & 57.8 & \multicolumn{1}{c}{ 66.3 \bfseries\imprv{1.0}} & \multicolumn{1}{c}{ 63.2 \bfseries\imprv{2.4}} \\
\midrule
\rowcolor{lightgray}
\multicolumn{10}{c}{close-sourced model} \\
\midrule
Gemini2.5Pro & \multicolumn{1}{c}{--} & 8 & 43.4 & 54.2 & 77.7 & 67.4 & 62.1 & 69.1 & 57.8 \\
\textbf{+ ours} & \multicolumn{1}{c}{--} & 8 &\multicolumn{1}{c}{\cellcolor{LavenderWash}71.6 \bfseries\imprv{28.2}}  & \multicolumn{1}{c}{ 56.6 \bfseries\imprv{2.4}} & 79.7 & 67.2 & 62.8 & \multicolumn{1}{c}{ 70.0 \bfseries\imprv{0.9}} & \multicolumn{1}{c}{62.2 \bfseries\imprv{4.4}} \\
\midrule
Gemini2.5Pro & \multicolumn{1}{c}{--} & 32 & 74.6 & 66.0 & 87.1 & 74.9 & 69.6 & 77.2 & 64.2 \\
\textbf{+ ours} & \multicolumn{1}{c}{--} & 32 &\multicolumn{1}{c}{ 80.9 \bfseries\imprv{6.3}} & \multicolumn{1}{c}{ 69.0 \bfseries\imprv{3.0}} & \multicolumn{1}{c}{87.1} & \multicolumn{1}{c}{76.1} & \multicolumn{1}{c}{70.9} & \multicolumn{1}{c}{78.0 \bfseries\imprv{0.8}} & \multicolumn{1}{c}{67.0 \bfseries\imprv{2.8}} \\
\midrule
GPT-4o & \multicolumn{1}{c}{--} & 8 & 58.3 & 55.38 & 67.2 & 58.6 & 53.5 & \multicolumn{1}{c}{59.7} & \multicolumn{1}{c}{49.4} \\
\textbf{+ ours} & \multicolumn{1}{c}{--} & 8 &\multicolumn{1}{c}{75.2 \bfseries\imprv{16.9}} & \multicolumn{1}{c}{60.5 \bfseries\imprv{5.1}} & 72.4 & 60.8 & 54.6 & \multicolumn{1}{c}{62.6 \bfseries\imprv{2.9}} &\multicolumn{1}{c}{\cellcolor{LavenderWash} 54.5 \bfseries\imprv{5.1}} \\
\midrule
GPT-4o & \multicolumn{1}{c}{--} & 32 & 71.3 & 59.6 & 69.3 & 61.1 & 55.8 & 62.1 & 49.9 \\
\textbf{+ ours} & \multicolumn{1}{c}{--} & 32 &\multicolumn{1}{c}{76.9 \bfseries\imprv{5.6}}  & \multicolumn{1}{c}{61.9 \bfseries\imprv{2.3}} & 70.6 & 62.7 & 54.8 & \multicolumn{1}{c}{62.7 \bfseries\imprv{0.6}} & \multicolumn{1}{c}{51.3 \bfseries\imprv{1.4}} \\
\bottomrule
\end{tabular}
\label{tab:main_results}
\end{table*}

\subsection{Experiment Setup}
\paragraph{Evaluation Benchmarks.}
We conduct experiments on three public benchmarks to evaluate our approach. Video-MME~\citep{fu2025video} comprises 900 videos and 2,700 multiple-choice Question-Answer pairs, categorized into three subsets based on video duration: short ($<$2 minutes), medium (4-15 minutes), and long (30-60 minutes). MLVU~\citep{zhou2025mlvu} includes videos ranging from 3 minutes to 2 hours and spans 9 tasks, with 2,174 multiple-choice VQA pairs. LVBench~\citep{wu2024longvideobench} features videos with an average duration of 4,101 seconds per video, which is the longest. It contains 1,549 multiple-choice VQA pairs across 6 tasks. Importantly, all datasets are human-annotated, ensuring high-quality labels for evaluation. To verify model-agnostic generality, we evaluated downstream tasks with open-source models, including Qwen2.5-VL-7B, Qwen2.5-VL-72B~\citep{qwen2.5vl} and Intern3.5-VL-8B~\citep{wang2025internvl3}; and closed-source models comprise ChatGPT-4o and Gemini 2.5 Pro~\citep{comanici2025gemini}.

\paragraph{Implementation Details.}

We train K-frames with Qwen2.5-VL-3B as the backbone. For each training and evaluation instance, we uniformly sample $T=256$ frames per video as inputs. K-frames predict continuous index ranges $[s,e]$ as highlight clips together with rationales. These key clips then guide the selection of k keyframes for the downstream task. Specifically, when the number of keyframes set to $k=8$, we employ Focused Sampling (exclusively from key clips). When $k=32/64$, we employ Hybrid Sampling (details in Appendix~\ref{hybrid}). As described in Sec. 3.1, we construct PeakClips for our Supervised Fine-Tuning, and randomly select 20K samples from  original LLaVA-Video-178K for our Reinforcement Learning optimization. During all 3 stages training, we freeze the vision encoder and update only the multimodal projector and the LLM. The first two supervised phase takes 36\,hours, and the RL phase 40\,hours. We use a learning rate of \(1.0\times10^{-5}\) for both supervised phases, and \(1.0\times10^{-6}\) for RL with a KL penalty coefficient of 0.01.


\subsection{Performance across General Video Benchmarks}

\paragraph{Temporal Localization.}
We first evaluate our K-frames on the Needle QA (a subset of MLVU). It constructs each example by randomly inserting a short ``needle" segment containing evidence into a longer background video and annotating a corresponding pair of questions and answers, thereby directly probing temporal grounding.
As shown in Table~\ref{tab:main_results}, compared to uniform sampling, our K-frames significantly improves performance on this task. For example, when using the Gemini 2.5 Pro as the downstream model with the number of frames set to $k=8$, our method boosts the accuracy from \textbf{43.4\%} to \textbf{71.6\%}, achieving a notable improvement of \textbf{28.2\%}. This is because our model can effectively align visual evidence to time span and then locate the relevant scenes.

\paragraph{Quantitative Analysis.}


\begin{figure*}[!t]
\centering
\includegraphics[width=1\textwidth]{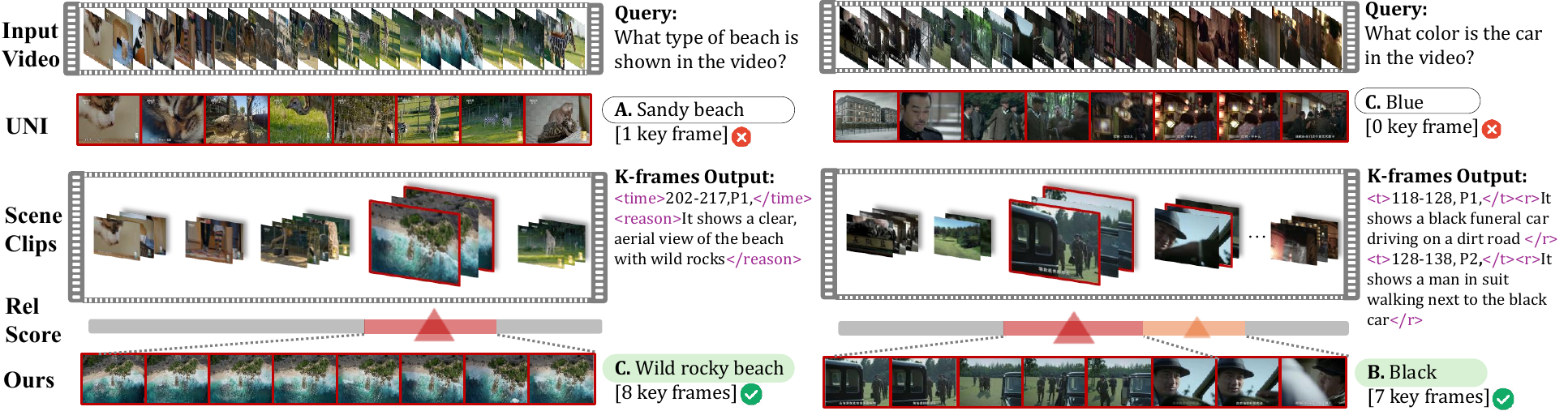} 
\caption{Qualitative comparison between uniform sampling and our K-frames method.}
\label{fig:cmp}
\end{figure*}

\begin{wrapfigure}{r}{0.45\textwidth} 
    \centering
    \small
    \setlength{\tabcolsep}{2.5pt}
    \captionof{table}{Comparison with ViaRL using the same backbone for frame selection.}
    \begin{tabular}{@{\hspace{\tabcolsep}} c c c c c @{\hspace{\tabcolsep}}}
        \toprule
        \multirow{2}{*}{Model} & \multirow{2}{*}{Size}& {\multirow{2}{*}{Frames}} & \multicolumn{2}{c}{MLVU}  \\
        \cmidrule(lr){4-5}
        & & & {N-Needle} & {M-Avg}  \\
        \midrule
        QwenVL2.5 & 7B & 8 & 58.6 & 53.9  \\
        ViaRL & 7B & 8 & 73.5 & 58.2  \\
        \rowcolor{gray!20}
        \textbf{ours} & 7B & 8 & \textbf{77.5} & \textbf{60.4} \\
        \bottomrule
    \end{tabular}
    \label{tab:comparation_results}
\end{wrapfigure}

We evaluate K-frames on several challenging long-video benchmarks. 
As shown in Table~\ref{tab:main_results}, our method consistently and significantly exceeds the baseline performance in different open-source and closed-source models. For example, when applied to the open source QwenVL-2.5-7B with $k=8$ frames, our approach achieves a improvement of \textbf{6.5\%} on MLVU (M- Avg). Similarly, when integrated with the closed-source GPT-4o with $k=8$, it boosts the LVBench score by a significant \textbf{5.1\%}. This is because our model can effectively localize relevant scenes by aligning visual evidence with its correct time span, enabling it to effectively extract keyframes.

Furthermore, our method demonstrates robust performance gains even as the number of sampled keyframes increases. Taking the Qwen2.5-VL-72B's performance on LVBench as an example, the baseline score scales from 55.6 with 8 frames to 59.9 with 64 frames. Our method also improves upon these scores at each step---achieving 59.3 (\textbf{+3.7}) and 61.1 (\textbf{+1.2}) respectively. This demonstrates the scalability and effectiveness of our K-frames across different sampling densities.


Using the same QwenVL2.5-3B backbone for frame selection, we compare our method against ViaRL~\citep{viarl}. It requires an iterative update strategy that involves jointly optimizing the downstream QwenVL2.5-7B model. In contrast, our method is truly plug-and-play, eliminating the need for costly downstream model optimization. As illustrated in Table~\ref{tab:comparation_results}, our approach outperforms ViaRL by \textbf{4.0} points on Needle-QA and \textbf{2.2} points on the MLVU M-Avg score. Moreover, unlike ViaRL's fixed 8-frame selection, our method can select any-k keyframes, which showcases the enhanced flexibility and superiority of our clip2frame paradigm. This paragraph continues to demonstrate how the text naturally wraps around the figure that has been placed on the right-hand side of the page.

\paragraph{Qualitative Analysis.}
Figure~\ref{fig:cmp} presents a qualitative analysis that visually contrasts the performance of K-frames against the widely used uniform sampling baseline. For instance, when asked to identify the car's color in the video, uniform sampling method is blind to the query, selecting frames from various irrelevant scenes. In contrast, K-frames showcases a more sophisticated understanding. Its relevance score identifies two distinct but semantically related scene clips: one showing the ``black funeral car driving" and another showing a ``man in suit walking next to the black car", which provides the downstream model with comprehensive and unambiguous visual evidence.
These examples demonstrate how K-frames successfully identifies and leverages critical visual evidence and leads to more accurate and well-grounded video understanding.

\subsection{Ablation Study}

\begin{table}[ht]

\begin{minipage}[t]{0.48\linewidth}
    \centering
    \small
    \setlength{\tabcolsep}{3pt}
    \captionof{table}{Ablation on training stages. Baseline is uniformly sampling. Downstream MLLM is Qwen2.5-VL-7B with $k=32$ frames.}
    \begin{tabular}{cccc S[table-format=2.1] S[table-format=2.1]}
        \toprule
        \textbf{MODEL} & \textbf{SFT 1} & \textbf{SFT 2} & \textbf{RL} & {\textbf{Needle-QA}} & {\textbf{MLVU}}\\
        \midrule
        baseline & -       & -       & -       & 63.4  & 61.7 \\
        SFT      & -       & \ding{51} & -       & 75.8 & 64.1 \\
        SFT      & \ding{51} & \ding{51} & -       & 76.3 & 64.5  \\
        RL       & \ding{51} & \ding{51} & \ding{51} & \textbf{79.4}  & \textbf{65.9} \\
        \bottomrule
    \end{tabular}
    \label{tab:ablation_stages}
\end{minipage}
\hfill 
\begin{minipage}[t]{0.48\linewidth}
    \centering
    \small
    \setlength{\tabcolsep}{4.5pt}
    \captionof{table}{Ablation on temporal prompts, performed on SFT2 model for efficient validation. The baseline uses a uniform sampling strategy.}
    \begin{tabular}{ccc S[table-format=2.1] S[table-format=2.1]}
        \toprule
        \textbf{MODEL} & $\mathbf{TP}$ & $\mathbf{VP}$ & {\textbf{Needle-QA}} & {\textbf{MLVU}}\\
        \midrule
        baseline & - & - & 63.4 & 61.7  \\
        SFT2 & \ding{51} & -       & 75.5 & 63.9  \\
        SFT2 & -         & \ding{51} & 70.4 & 62.2  \\
        SFT2 & \ding{51} & \ding{51} & \textbf{75.8} & \textbf{64.1}  \\
        \bottomrule
    \end{tabular}
    \label{tab:ablation_prompts}
\end{minipage}
\vspace{-2ex}
\end{table}

\paragraph {Ablation on Training Stages.} We first analyze the contribution of each stage in our multi-stage progressive curriculum. As shown in Table~\ref{tab:ablation_stages}, training with only the second SFT stage yields a  score of 64.1 on MLVU. SFT2 is a necessary course because the model learns to predict query-conditioned key clips during SFT2, making it a prerequisite for the keyframe selection. SFT1 is a preliminary curriculum focused on foundational skills. Incorporating SFT1 provides a further gain, reaching 64.5 on MLVU and 76.3 on Needle-QA, which demonstrates this phase enhanced K-frames temporal grounding, which helps to final secene-driven keyframe selection. Moreover, adding the final Reinforcement Learning (RL) stage achieves a significant improvement, boosting performance by \textbf{1.4}\% on MLVU and \textbf{3.1}\% on Needle-QA. This is because RL stage directly optimizes its clip2frame selection policy to align with the downstream tasks.

\paragraph{Ablation on Different Temporal Prompts.} Given the importance of temporal cues, we next explore the individual contributions of our two temporal prompts: Visual Prompt (VP) and Textual Prompt (TP). As shown in Table~\ref{tab:ablation_prompts}, using only VP or TP results in suboptimal performance. This limitation suggests that relying on a single type of prompt provides an incomplete representation of the video's temporal structure. In contrast, combining both VP and TP attains a final score of \textbf{64.1}\% on MLVU. It is because our two prompts capture complementary information. The VP directly provides visual evidence, while the TP offers fine-grained, position-specific guidance for each frame. This synergy allows K-frames to build a more comprehensive understanding of temporal dynamics, enhancing its scene-driven keyframe selection performance.



    
    


\section{Conclusion}
In this work, we introduce K-frames, a new scene-driven paradigm for long-video understanding. It reframes keyframe selection as a clip2frame prediction task, preserving scene continuity while enabling flexible any-k sampling. To realize this paradigm, we first construct PeakClips, a new 200K query-clip relevance dataset. We then propose a three-stage SFT-RL training framework designed to produce a powerful key clip predictor that is highly optimized for downstream tasks. Extensive experiments show K-frames acts as an effective, interpretable, and model-agnostic front-end, consistently boosting MLLM performance on major long-video benchmarks.

\bibliography{iclr2026_conference}
\bibliographystyle{iclr2026_conference}

\newpage
\appendix
\section{Use of Large Language Models (LLMs)}
During the preparation of this manuscript, we utilized a Large Language Model (LLM) as a writing assistant. The primary application of the LLM was for language enhancement, which included improving grammar, refining wording for conciseness, and rephrasing sentences to improve clarity and flow.

In accordance with the established ethical guidelines, we, the authors, affirm that we are fully responsible for the content of this submission. All text, including any passages refined with the assistance of the LLM, has been critically reviewed, edited, and validated by the authors. The scientific claims, results, and conclusions presented herein are our own. We are solely responsible for any potential errors, inaccuracies, or ethical violations in this work.

\section{Comparison with Prior Work}

\begin{figure*}[!htbp]
\centering
\includegraphics[width=1\textwidth]{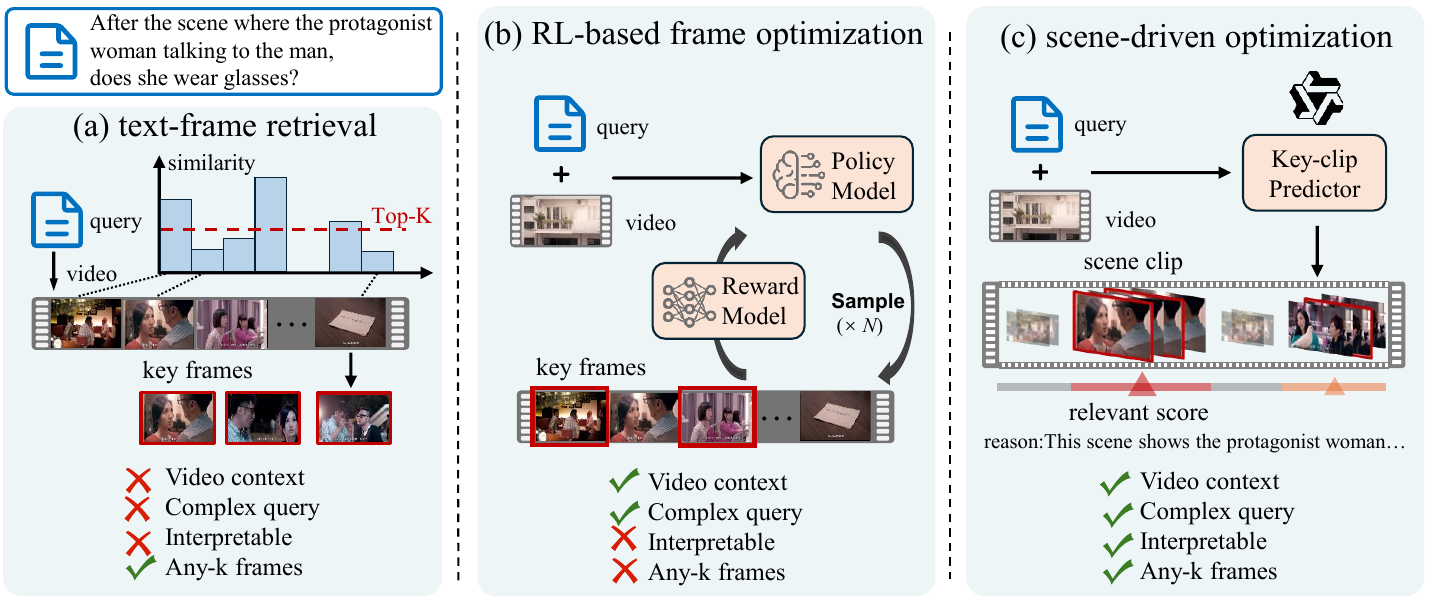}
\caption{Comparison with existing keyframe selection methods.}
\label{fig:related}
\end{figure*}
\vspace{-2ex}

As illustrated in Figure~\ref{fig:related}, our scene-driven K-frames paradigm addresses the main limitations of existing approaches. Text–frame retrieval methods treat videos as independent frame sets and rank them by query similarity, overlooking temporal context and offering limited interpretability. RL-based frame optimization considers selection as a combinatorial search that often yields a sparse, disconnected set of frames, breaking scene continuity and degrading downstream performance. They are also typically tuned for a fixed number of frames, lacking flexibility for any-k selection and making it difficult to meet personalized compute budgets. In contrast, our K-frames predicts semantically coherent, query-relevant clips and then samples keyframes, inherently preserving temporal continuity and providing interpretable clip-level rationales. Moreover, it supports flexible any-k selection, allowing users to balance performance and computational cost.

\section{Dataset Construction and Analysis}
\subsection{Implementation Details}
In this section, we present how we organize our prompt to generate labels using LLM.

\paragraph{Caption Generation.}
To obtain fine-grained scene-level descriptions after video segmentation, we employed an instruction-following style prompt, in which the model is explicitly assigned the role of a Professional Video Content Analyst. The prompt enforces a strict JSON output format containing three components: \texttt{scenes}, \texttt{chapters}, and \texttt{video\_summary}. 

\begin{figure*}[!h]
\centering
\includegraphics[width=1\textwidth]{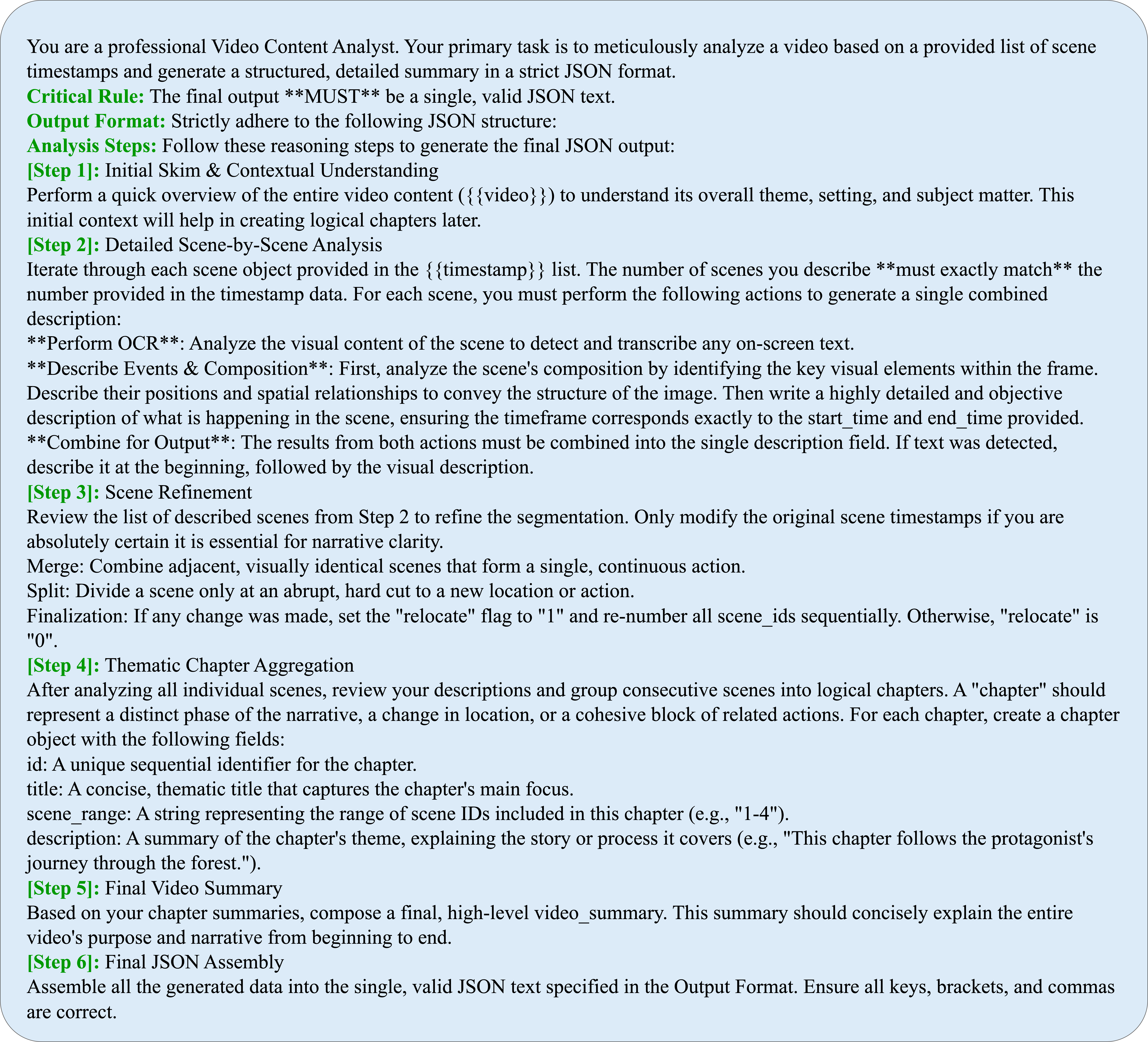} %
\caption{Prompt used for generating scene-level captions.}
\label{fig:prompt1}
\end{figure*}

As shown in Figure~\ref{fig:prompt1}, our Instructional Prompt for caption generation is designed to guide the LLM through a structured, multi-stage analysis. The prompt instructs the model to first perform an initial skim for overall context, followed by a detailed scene-by-scene analysis that combines OCR of on-screen text with a compositional description of visual elements . Subsequent instructions direct the model to refine scene boundaries by merging or splitting segments, aggregate related scenes into thematic chapters, and conclude with a high-level video summary . This step-wise, instruction-based format enforces a highly structured analytical process, resulting in objective and detailed video descriptions suitable for our dataset.



\paragraph{Relevance Scoring.}
To evaluate the relevance of each scene in the context of video question answering (VideoQA), we employed a second evaluation-oriented Instrutional Prompt, positioning the model as a Video QA Relevance Analyst. The output is again required to follow a strict JSON structure, including the fields \texttt{scene\_id}, \texttt{relevance\_score}, and \texttt{reason}.  

\begin{figure*}[!h]
\centering
\includegraphics[width=1\textwidth]{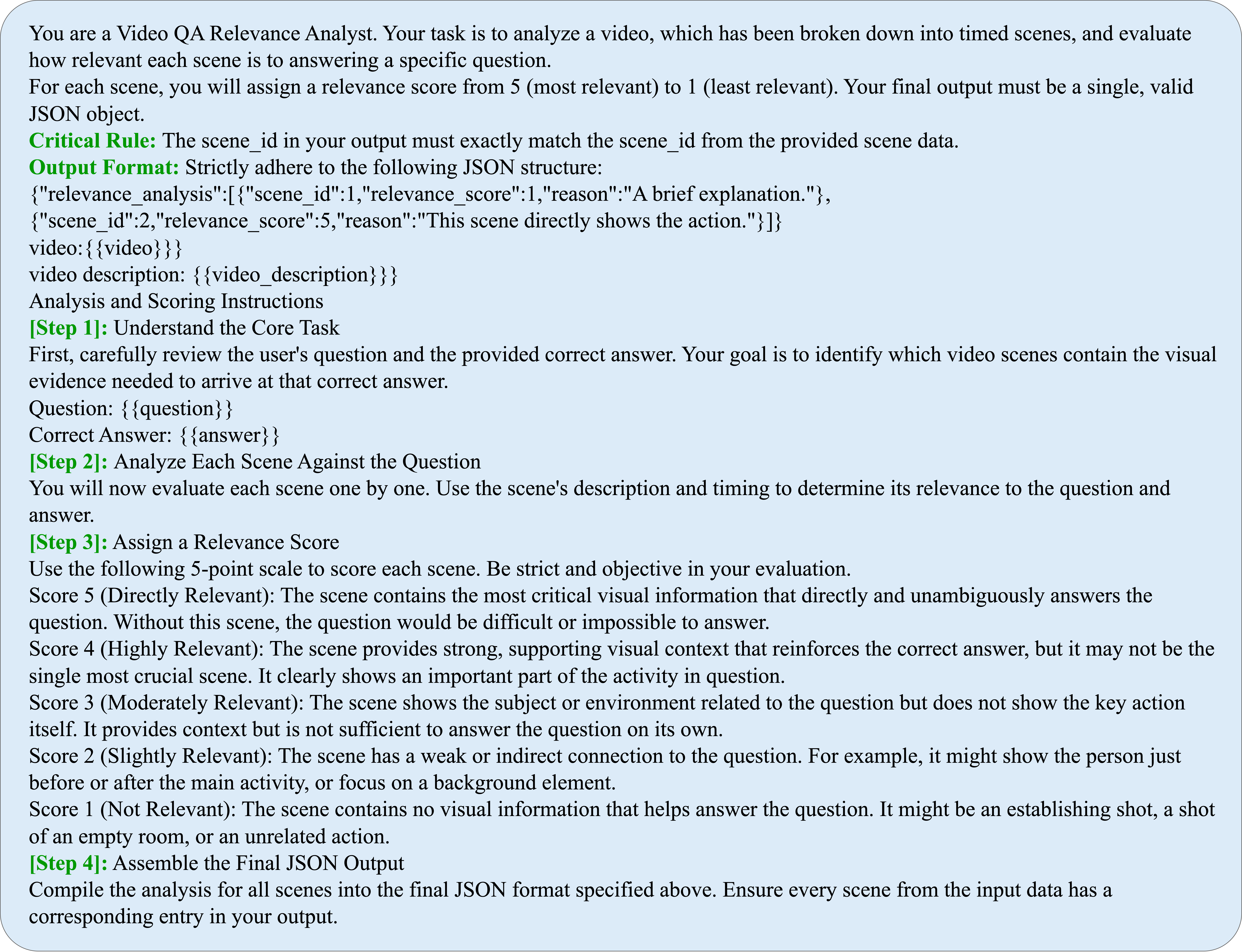} %
\caption{Prompt used for generating scene-level query relevance scores.}
\label{fig:prompt2}
\end{figure*}

As illustrated in Figure \ref{fig:prompt2}, the procedure begins by providing the model with the question and the corresponding gold-standard answer, which serve as the reference criteria. Each scene is then assessed with respect to its contribution toward answering the question. Relevance is assigned according to a five-point ordinal scale:  
\begin{itemize}
    \item \textbf{5 (Directly Relevant)}: the scene contains critical visual evidence that directly resolves the question;  
    \item \textbf{4 (Highly Relevant)}: the scene provides strong supporting context, though it is not the single most essential frame;  
    \item \textbf{3 (Moderately Relevant)}: the scene depicts related subjects or environments but lacks the decisive information;  
    \item \textbf{2 (Slightly Relevant)}: the scene has only weak or indirect connection to the question;  
    \item \textbf{1 (Not Relevant)}: the scene provides no information useful for answering the question.  
\end{itemize}

Each score must be accompanied by a concise justification (\texttt{reason}), ensuring interpretability and consistency across all annotations. This prompt design enforces rigorous evaluation criteria, quantitative scoring, and machine-readable outputs that are suitable for large-scale automated processing.

\subsection{Dataset Statistics}
\label{appendix:dataset}
In this section, we present additional statistical details of the \textbf{PeakClips} dataset. We randomly sampled 6702 videos from LLAVA-Video-178K and adopted Gemini for the labeling. As listed in Table \ref{tab:peakclips_stats}, the PeakClips dataset comprises over 200k annotations in total, including 6,702 annotated videos, 108,221 scenes with 281,643 corresponding relevance scores, and 19,070 chapters, with an average of 16.15 scenes and 2.85 chapters per video. Since the PeakClips dataset is derived from four sources—NextQA, Academic, YouTube, and PerceptionTest—we present in Table \ref{tab:source_stats} the number of videos, scenes, and relevance scores associated with each source.

\begin{table}[h!]
\centering
\caption{Annotation statistics of the PeakClips dataset.}
\begin{tabular}{lcc}
\toprule
\textbf{Annotation Type} & \textbf{Count} & \textbf{Average per Video} \\
\midrule
Video-level Summarization        & 6,702   & 1    \\
Chapter-level Description        & 19,070  & 2.85  \\
Scene-level Description          & 108,221 & 16.15 \\
Relevance Query                  & 16,883  & 2.52 \\
Scene-level Relevance Scores     & 281,643 & 42.02 \\
\midrule
\textbf{Total Annotations} & \multicolumn{2}{c}{281,643} \\
\bottomrule
\end{tabular}
\label{tab:peakclips_stats}
\end{table}

\begin{table}[h!]
\centering
\caption{Scene relevance score statistics of PeakClips dataset across sources.}
\resizebox{\textwidth}{!}{%
\begin{tabular}{lrrrrrrrr}
\toprule
\textbf{Source} & \textbf{Videos} & \textbf{Scenes} & \textbf{Score 1} & \textbf{Score 2} & \textbf{Score 3} & \textbf{Score 4} & \textbf{Score 5} & \textbf{Total} \\
\midrule
\textbf{Global}       & 6,702  & 281,643 & 88,588 & 67,842 & 28,154 & 36,220 & 36,263 & 281,643 \\
\midrule
NextQA                & 716   & 10,444  & 2,004  & 2,460  & 1,453  & 1,713  & 1,648  & 10,444 \\
Academic              & 1,512  & 72,961  & 17,791 & 16,168 & 8,436  & 11,676 & 11,817 & 72,961 \\
YouTube               & 4,086  & 193,976 & 68,594 & 48,833 & 17,816 & 21,903 & 21,624 & 193,976 \\
PerceptionTest        & 388   & 4,262   & 199    & 381    & 449    & 928    & 1,174   & 4,262 \\
\bottomrule
\end{tabular}%
}
\label{tab:source_stats}
\end{table}

\begin{figure*}[!h]
\centering
\includegraphics[width=1\textwidth]{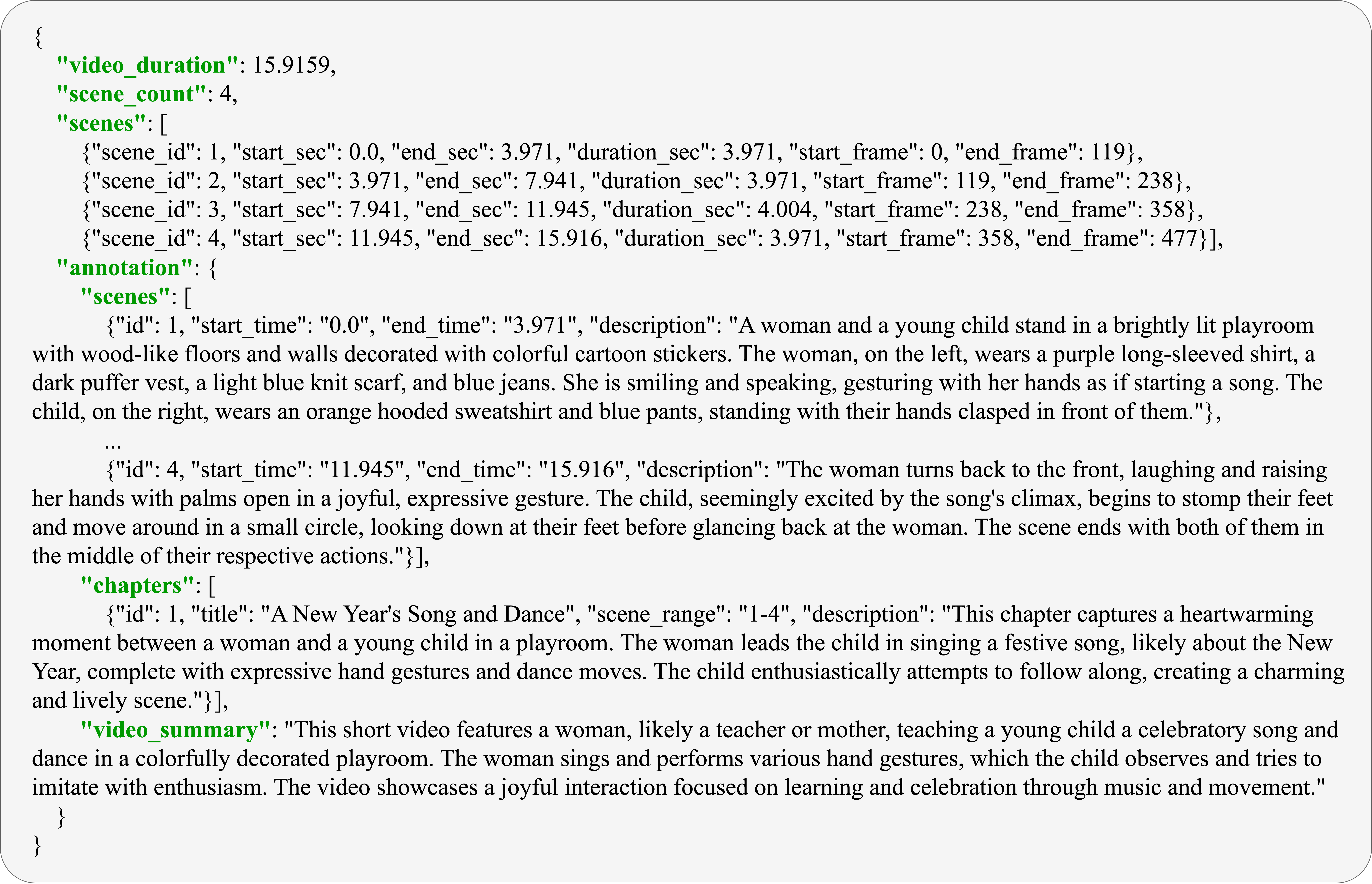} %
\caption{An example data of Scene/Chapter/Video-level annotation by LLM in PeakClips.}
\label{fig:data1}
\end{figure*}

\begin{figure*}[!h]
\centering
\includegraphics[width=1\textwidth]{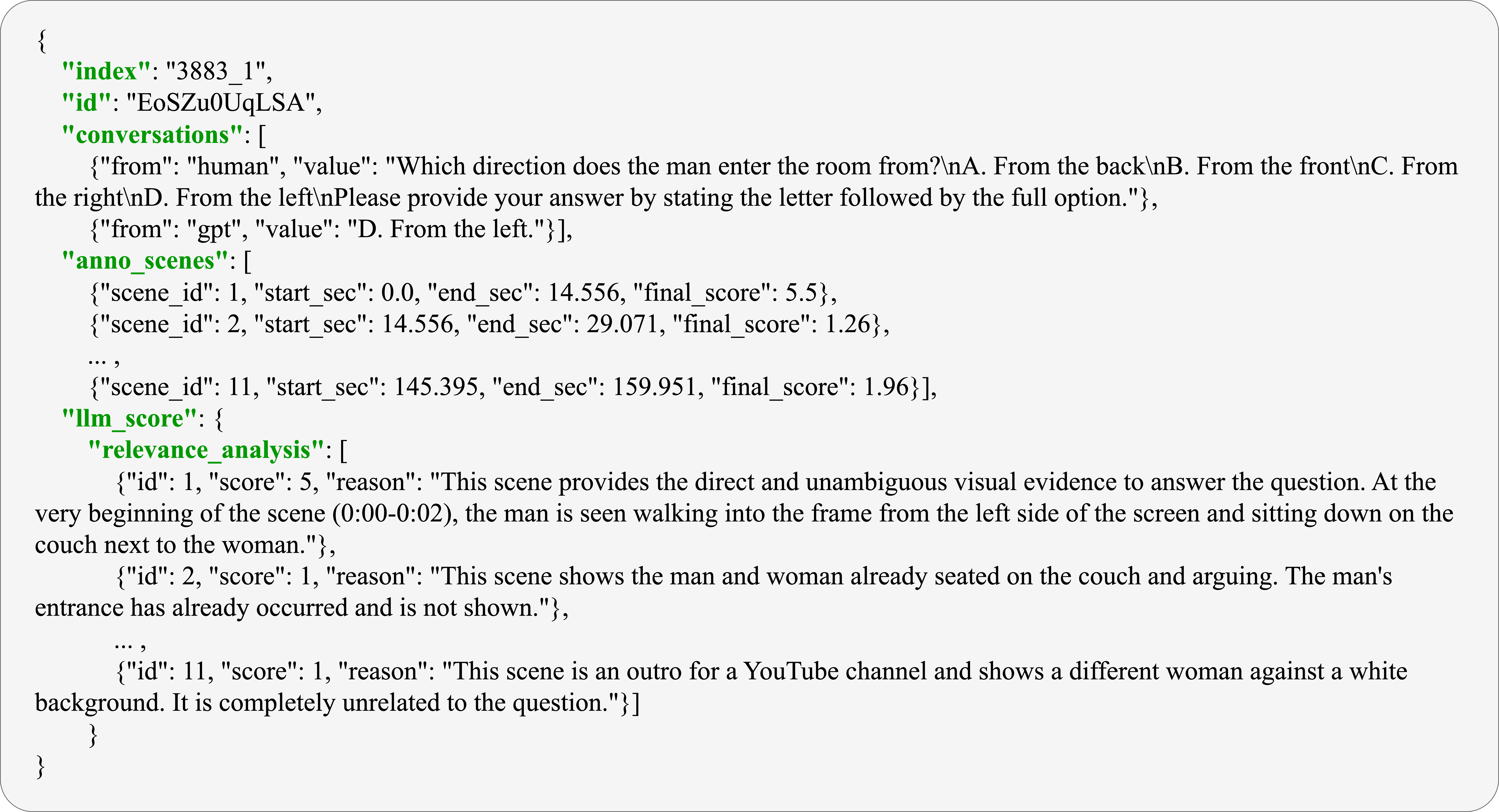} %
\caption{An example data of scene relevance score labeled by LLM in PeakClips.}
\label{fig:data2}
\end{figure*}

Below are two sample entries from the \textbf{PeakClips} dataset, illustrating (i) scene/chapter/video-level annotations (Figure \ref{fig:data1}) and (ii) scene--query relevance annotations (Figure \ref{fig:data2}). 

\vspace{-2ex}

\section{Training Details}

\subsection{Instructional Prompts}
\lstset{
  basicstyle=\ttfamily\small, 
  breaklines=true,            
  showstringspaces=false,     
  frame=single,               
  caption={Instructional Prompt}, 
  label={lst:my_prompt}       
}

\label{sec:appendix_prompt}

This section provides the detailed instructional prompts used during the different stages of our training curriculum for K-frames.

\paragraph{Instructional Prompts for SFT1.}
In the first SFT stage, we employ three task-specific prompts to instill foundational temporal grounding capabilities in the model. Each prompt is designed to teach a core sub-task.

\text{Caption-to-Scene Localization.}
This task trains the model to identify the temporal boundaries (start and end frames) of a scene given its textual description. The prompt used is:

\begin{figure*}[!h]
\centering
\includegraphics[width=1\textwidth]{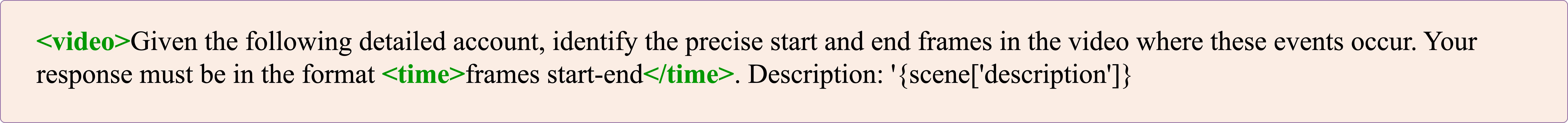} %
\caption{Instructional prompt for Caption-to-Scene Localization.}
\label{fig:fig11}
\end{figure*}
\vspace{-2ex}

Scene-to-Caption Generation.
As a dual task, this prompt instructs the model to generate a concise and accurate description for a given temporal segment of the video. The prompt used is:

\begin{figure*}[!h]
\centering
\includegraphics[width=1\textwidth]{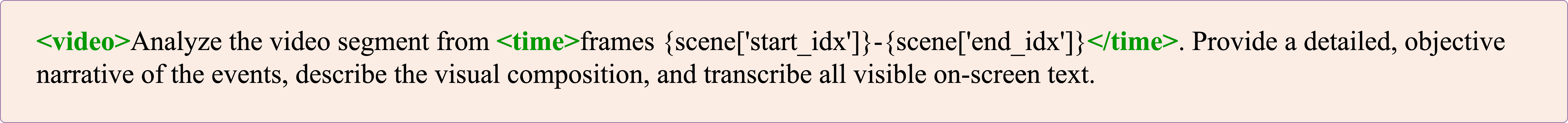} %
\caption{Instructional prompt for Scene-to-Caption Generation.}
\label{fig:fig12}
\end{figure*}
\vspace{-2ex}

Clip-Query Relevance Scoring.
This task requires the model to assess and score the relevance of a specific video clip in relation to a given query, helping it learn to weigh the importance of different scenes. The prompt used is:

\begin{figure*}[!h]
\centering
\includegraphics[width=1\textwidth]{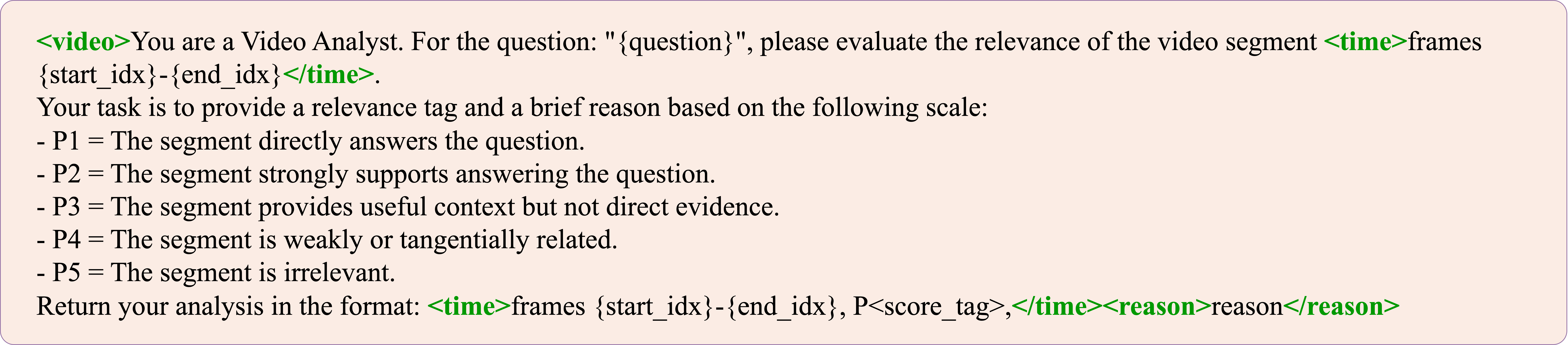} %
\caption{Instructional prompt for Clip-Query Relevance Scoring.}
\label{fig:fig13}
\end{figure*}
\vspace{-2ex}

\paragraph{Instructional Prompts for SFT2 and RL.}
The second SFT stage uses a comprehensive prompt to train the model for its primary goal: predicting a complete set of highlight clips for a given video and query. This same prompt is then used by the actor model during the Reinforcement Learning (RL) stage to generate actions (i.e., predict key clips).

The prompt instructs the model to identify all relevant clips, assign a priority level (\textbf{P1} or \textbf{P2}) to each, and provide a brief rationale for its selection. The prompt used is:

\begin{figure*}[!h]
\centering
\includegraphics[width=1\textwidth]{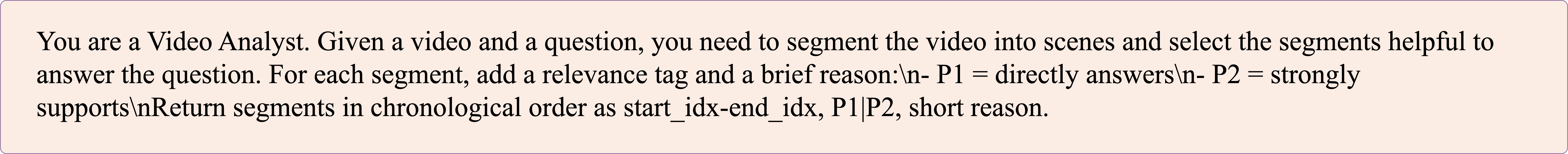} %
\caption{Instructional prompt for Clip-Query Relevance Scoring.}
\label{fig:fig14}
\end{figure*}

\subsection{Inference details}
After predicting the initial set of query-relevant key clips, a subsequent step is required to select the final $k$ keyframes. This section details the two methodologies we propose for this task: \textbf{Focused Sampling}, which selects keyframes exclusively from the predicted clips, and \textbf{Hybrid Sampling}, which dynamically samples from both the predicted clips and the background regions.
\label{appd:infer}
\paragraph{Focused Sampling.}
Given a video frame sequence $\{f_1,\ldots,f_T\}$, K-frames first predicts a set of
query-relevant key clips
$\vc=\{([a_j,b_j],\, t_j)\}_{j=1}^{M}$,
where $t_j\in\{\mathrm{P1},\mathrm{P2}\}$ denotes the importance type and
$\ell_j=b_j-a_j+1$ is the length.
We select $k$ keyframes \emph{exclusively} from these predicted clips.
Let $(\wPone,\wPtwo)=(2,1)$ be class weights (P1 is twice as important as P2).
We allocate the per-clip budget by weighted proportion:
\begin{equation}
\label{eq:focused_alloc}
k_j \;=\; \operatorname{round}\!\left(
K \cdot \frac{w(t_j)\,\ell_j}{\sum_{i=1}^{M} w(t_i)\,\ell_i}
\right),
\qquad
\begin{cases}
w_{\mathrm{P1}}, & t=\mathrm{P1},\\[2pt]
w_{\mathrm{P2}}, & t=\mathrm{P2}.
\end{cases}
\end{equation}
To prevent short P1 clips from receiving zero frames, we enforce a \emph{P1-at-least-1}
guarantee by borrowing from donors with $k_i>1$ (prefer P2 donors); if the global budget
is insufficient, the guarantee is relaxed.
Inside each clip, we sample \emph{Uniformly}: pick $k_j$ equally spaced frames from
$\{f_{a_j},\ldots,f_{b_j}\}$ under a chronological constraint (strictly increasing frame
indices across clips). If the total selected frames are fewer than $k$ due to rounding or
chronology constraints, we top up uniformly from the non-key region after the last picked
index.


\begin{algorithm}[h]
\caption{Focused Sampling}
\label{alg:focused}
\textbf{Require} Predicted clips $\vc = \{([a_j,b_j],t_j)\}_{j=1}^M$, target $k$, 
weights $(\wPone,\wPtwo) = (2,1)$\;
Merge adjacent same-type clips within tolerance $\tau=2$ (reasons concatenated)\;
Compute $k_j$ by equation~\eqref{eq:focused_alloc}, fix rounding so that $\sum_j k_j = k$\;
Enforce P1-at-least-1 by borrowing from donors with $k_i>1$ (prefer P2 donors)\;
last\_id $\gets -\infty$\;
\ForEach{$j \gets 1$ \textbf{to} $M$}{
    $\mathcal{C}_j \gets \{f_{a_j},\ldots,f_{b_j}\}$ restricted to frame id $>$ last\_id\;
    Select $k_j$ equally spaced frames from $\mathcal{C}_j$\;
    Update last\_id to the largest picked id\;
}
\If{selected $<K$}{
    Top up uniformly from non-key frames with id $>$ last\_id\;
}
Return $k$ frames sorted by index\;
\end{algorithm}

\begin{table}[h]
\centering
\caption{Focused Sampling hyperparameters.}
\label{tab:focused_hparams}
\begin{tabular}{@{}ll@{}}
\toprule
Merge tolerance $\tau$ & $2$ \\
Segment weights & $w_{\mathrm{P1}}=2,\ w_{\mathrm{P2}}=1$ \\
P1 guarantee & at least one frame for P1 if budget allows \\
Chronology constraint & strictly increasing frame indices \\
\bottomrule
\end{tabular}
\end{table}

\paragraph{Hybrid Sampling.}
\label{hybrid}
We partition all candidate frames $F$ into predicted frames $\vp$ (inside key clips)
and background frames $\vb$ (the rest). We first allocate a global share between $\vp$
and $\vb$, then distribute the predicted share across clips as in
Focused Sampling (uniform only). Let $\apred$ be the prediction-to-background length
weight ($\apred:1 = 4:1$ in our default) and let $\rmin\in[0,1]$ be a lower bound on
the predicted share (e.g., $\rmin=0.5$).
With $|\vp|$ and $|\vb|$ the available counts, we compute
\begin{align}
\label{eq:hybrid_raw}
k_{\vp}^{\mathrm{raw}}
&= \operatorname{round}\!\left(
K \cdot \frac{\apred\,|\vp|}{\apred\,|\vp| + |\vb|}
\right),\\
\label{eq:hybrid_kp}
k_{\vp}
&= \min\!\Big(\,|\vp|,\ \max\big(\lceil K\rmin\rceil,\, k_{\vp}^{\mathrm{raw}}\big)\Big),\\
\label{eq:hybrid_kb}
k_{\vb}
&= \min\!\big(|\vb|,\ K-k_{\vp}\big).
\end{align}
If $k_{\vp}+k_{\vb}<k$ due to upper caps, remaining slots are assigned to the side that
still has capacity. Inside $\vp$, we further allocate $k_{\vp}$ across clips using
\eqref{eq:focused_alloc} with the P1-at-least-1 guarantee, and select frames uniformly
within each clip. From $\vb$, we select $k_{\vb}$ frames uniformly. The final set is
deduplicated and sorted by frame index.

\begin{algorithm}[h]
\caption{Hybrid Sampling}
\label{alg:hybrid}
\textbf{Require} Predicted clips $\vs$, full candidates $F$, target $k$,
weight $\apred=4$, minimum ratio $\rmin$ (e.g., $0.5$)\;
Build a mask from $\vs$ to partition $F$ into $\vp$ and $\vb$\;
Compute $k_{\vp}$ by~\eqref{eq:hybrid_raw}--\eqref{eq:hybrid_kp}; set $k_{\vb}=k-k_{\vp}$ and cap by availability; top up if needed\;
Distribute $k_{\vp}$ across clips via~\eqref{eq:focused_alloc} with P1-at-least-1\;
Uniformly select frames within each predicted clip to meet its allocation\;
Uniformly select $k_{\vb}$ frames from $\vb$; union, deduplicate, sort\;
\textbf{Return} $k$ frames.
\end{algorithm}

\begin{table}[h]
\centering
\caption{Hybrid Sampling hyperparameters and defaults.}
\label{tab:hybrid_hparams}
\begin{tabular}{@{}ll@{}}
\toprule
Pred:background weight $\apred:1$ & $4:1$ \\
Minimum predicted ratio $\rmin$ & $0.5$ (configurable) \\
Within-pred clip weights & P1:2, P2:1; P1-at-least-1 guarantee \\
Chronology constraint & strictly increasing frame indices \\
\bottomrule
\end{tabular}
\end{table}

\section{More Experimental Results}

In this section, we provide additional quantitative and qualitative experimental results on long-video understanding benchmarks to further validate the effectiveness and generalizability of our proposed keyframe selection method. 

\subsection{More Results on long-video benchmark}

\begin{table*}[h]
\centering
\small
\setlength{\tabcolsep}{3.5pt}
\setlength{\aboverulesep}{0pt}
\arrayrulecolor{black}
\caption{More results on long-video understanding benchmarks. The red text indicates the performance improvement over the baseline (uniform sampling).}
\begin{tabular}{@{\hspace{\tabcolsep}} c c c c c c c c c c @{\hspace{\tabcolsep}}}
\toprule
\multirow{2}{*}{Models} & \multirow{2}{*}{Size}& {\multirow{2}{*}{Frames}} & \multicolumn{2}{c}{MLVU} & \multicolumn{4}{c}{VideoMME} & {\multirow{2}{*}{LVBench}}\\
\cmidrule(lr){4-5}
\cmidrule(lr){6-9}
& & & {Needle-QA} & {M-Avg} & {Short} & {Medium} & {Long} & {Avg} & \\
\midrule
\midrule
InternVL3.5 & 8B & 8 & 60.3 & 60.5 & 68.0 & 56.7 & 49.7 & 58.1 & 57.7\\
\textbf{+ ours} & 8B & 8 & \multicolumn{1}{c}{ 72.7 \bfseries\imprv{12.4}} & \multicolumn{1}{c}{ 60.4 \bfseries\imprv{6.5}} & 71.4 & 59.0 & 50.7 & \multicolumn{1}{c}{60.4 \bfseries\imprv{2.3}} & \multicolumn{1}{c}{60.0 \bfseries\imprv{2.3}}\\
\midrule
InternVL3.5 & 8B & 32 & 72.4 & 67.0 & 75.7 & 64.3 & 53.9 & 64.6 & 60.1 \\
\textbf{+ ours} & 8B & 32 & \multicolumn{1}{c}{74.9 \bfseries\imprv{2.5}} & \multicolumn{1}{c}{68.4 \bfseries\imprv{1.4}} & 75.9 & 64.2 & 55.1 & \multicolumn{1}{c}{65.1 \bfseries\imprv{0.5}} &  \multicolumn{1}{c}{61.8 \bfseries\imprv{1.7}}\\
\bottomrule
\end{tabular}
\label{tab:internvl_results}
\end{table*}

As shown in Table~\ref{tab:internvl_results}, we further assess our method’s generalizability by pairing it with InternVL-3.5. The consistent gains show that our scene-driven keyframe selection paradigm provides a provides an effective, interpretable, and plug-and-play solution for long video understanding.

As illustrated in Figure~\ref{fig:radar_mlvu} and Figure~\ref{fig:radar_videomme}, We further provide detailed visualizations of the results showing the sub-category performance on the MLVU and VideoMME datasets evaluated with the Qwen2.5-VL 7B model using 8 input frames. These results show that our model consistently improves performance across different task types on the evaluation benchmarks, with particularly notable gains on the Needle-QA localization task in MLVU. This result underscores the core strength of our approach: by predicting query-relevant clips, K-frames  preserve the temporal continuity and focus on informative clips. 
Moreover, we observe no improvement in the Topic Reasoning task of MLVU and Information Synopsis task of VideoMME. This is likely because such tasks typically require a holistic understanding of the entire video to reach a conclusion. In these global-level queries, our method appropriately predict relevant content spans with broader temporal coverage, often encompassing nearly the entire video. As a result, the subsequent keyframe selection reduces to uniform sampling over the whole video, yielding comparable performance. This observation highlights a specific scenario where our approach converges with the baseline.


\begin{figure*}[!ht]
    \centering
    \begin{minipage}[t]{0.48\textwidth}
        \centering
        \includegraphics[width=\linewidth]{figures/mlvu.png}
        \captionof{figure}{Performance on some MLVU subtasks. The downstream model is Qwen2.5-VL-7B with frames $k=8$.}
        \label{fig:radar_mlvu}
    \end{minipage}
    \hspace{\fill} 
    \begin{minipage}[t]{0.48\textwidth}
        \centering
        \includegraphics[width=\linewidth]{figures/mme.png}
        \captionof{figure}{Performance on VideoMME subtasks. The downstream model is Qwen2.5-VL-7B with frames $k=8$.}
        \label{fig:radar_videomme}
    \end{minipage}
\end{figure*}



\subsection{More Ablation Analysis}

\begin{table*}[h]
\centering
\small
\caption{Ablation study on the utility of including K-frames' generated reason text in the downstream model's prompt. ``+ ours'' refers to using our keyframe selection method. ``+ ours*'' indicates that in addition to using our selected frames, the textual reason for each clip's selection was also included in the prompt for the downstream model.}
\label{tab:reason_results}
\setlength{\tabcolsep}{4.5pt}
\setlength{\aboverulesep}{0pt}
\arrayrulecolor{black}
\begin{tabular}{@{\hspace{\tabcolsep}} c c c c c c c c c @{\hspace{\tabcolsep}}}
\toprule
\multirow{2}{*}{Models} & \multirow{2}{*}{Size}& {\multirow{2}{*}{Frames}} & {\multirow{2}{*}{MLVU}} & \multicolumn{4}{c}{VideoMME} & {\multirow{2}{*}{LVBench}}\\
\cmidrule(lr){5-8}
& & & & {Short} & {Medium} & {Long} & {Avg} & \\
\midrule
InternVL3.5 & 8B & 8 & 60.5 & 68.0 & 56.7 & 49.7 & 58.1 & 57.7 \\
+ ours* & 8B & 8 & 65.6 & 67.7 & 57.8 & 47.9 & 57.8 & 52.5 \\
\rowcolor{lightgray}
\textbf{+ ours} & 8B & 8 & \textbf{64.4} & \textbf{71.4} & \textbf{59.0} & \textbf{50.7} & \textbf{60.4} & \textbf{60.0} \\
\midrule
InternVL3.5 & 8B & 32 & 67.0 & 75.7 & 64.3 & 53.9 & 64.6 & 60.1 \\
+ ours* & 8B & 32 & 65.3 & 68.0 & 58.3 & 46.9 & 57.7 & 53.8 \\
\rowcolor{lightgray}
\textbf{+ ours} & 8B & 32 & \textbf{68.4} & \textbf{75.9} & \textbf{64.2} & \textbf{55.1} & \textbf{65.1} & \textbf{61.8} \\
\bottomrule
\end{tabular}
\end{table*}

\paragraph{Ablation of Including Reason Text in Downstream Prompts.} We conducted an ablation study to determine whether the textual explanations generated by K-frames for clip selection could further improve downstream task performance. To do this, we appended the reason text to the prompt given to the final downstream QA model. The results, presented in Table~\ref{tab:reason_results}, shows: while our K-frames selection method (+ ours) significantly outperforms the baseline, including the reason text (+ ours*) degrades performance across most benchmarks.

We attribute the observed performance degradation to the design of K-frames. The selector is a lightweight MLLM whose core strength is relevance discrimination—identifying query-relevant segments—rather than accurate answer generation. Consequently, the accompanying reason text, although correctly indicating relevance, may include the selector’s own preliminary reasoning or partial answers. These artifacts can introduce distracting or misleading cues that interfere with the downstream model’s more sophisticated reasoning process, leading to reduced end-to-end performance.

\subsection{More Qualitative Analysis}

\begin{figure*}[!t]
\centering
\includegraphics[width=1\textwidth]{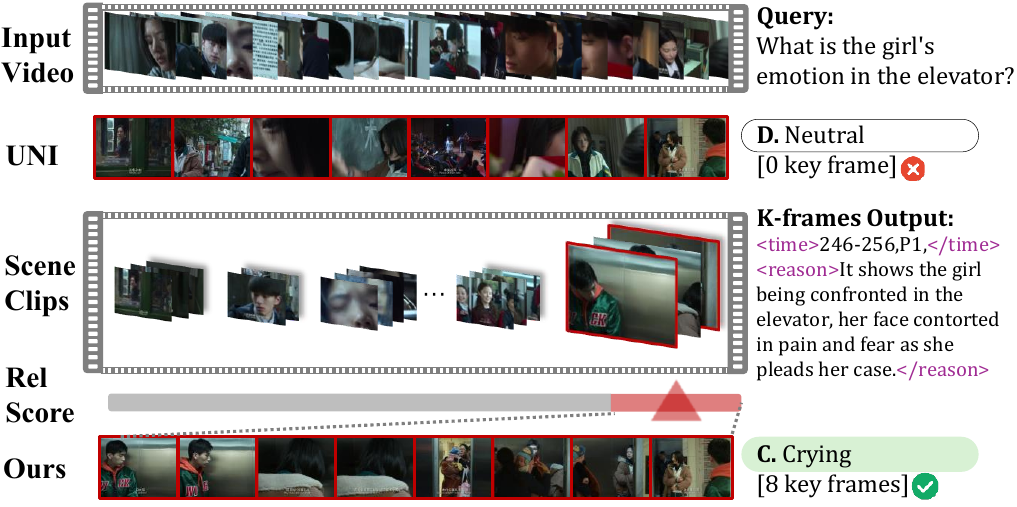} 
\caption{Qualitative comparison between uniform sampling and our K-frames method with the number of frames set to $k=8$.}
\label{fig:cmp_8_appendix}
\end{figure*}

\begin{figure*}[!t]
\centering
\includegraphics[width=1\textwidth]{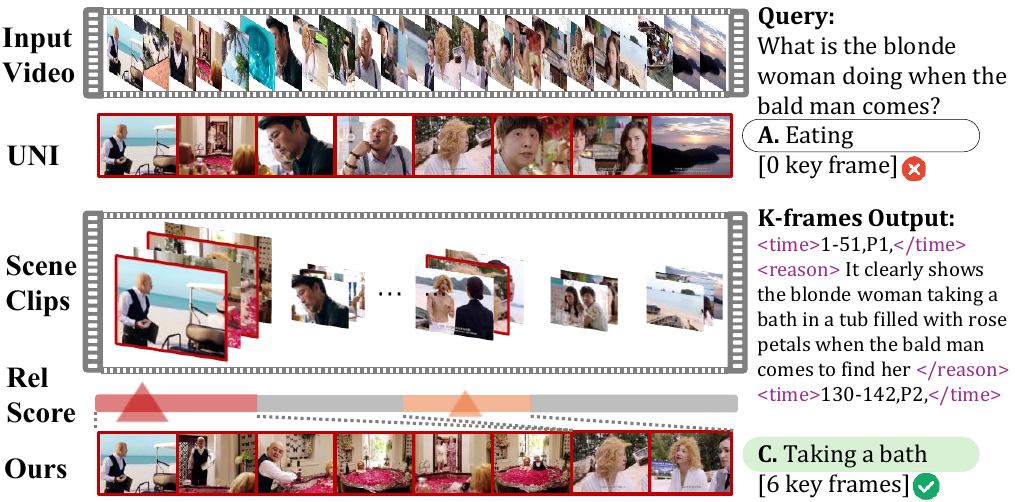} 
\caption{Qualitative comparison between uniform sampling and our K-frames method with the number of frames set to $k=8$.}
\label{fig:cmp_8_appendix_2}
\end{figure*}

\begin{figure*}[!t]
\centering
\includegraphics[width=1\textwidth]{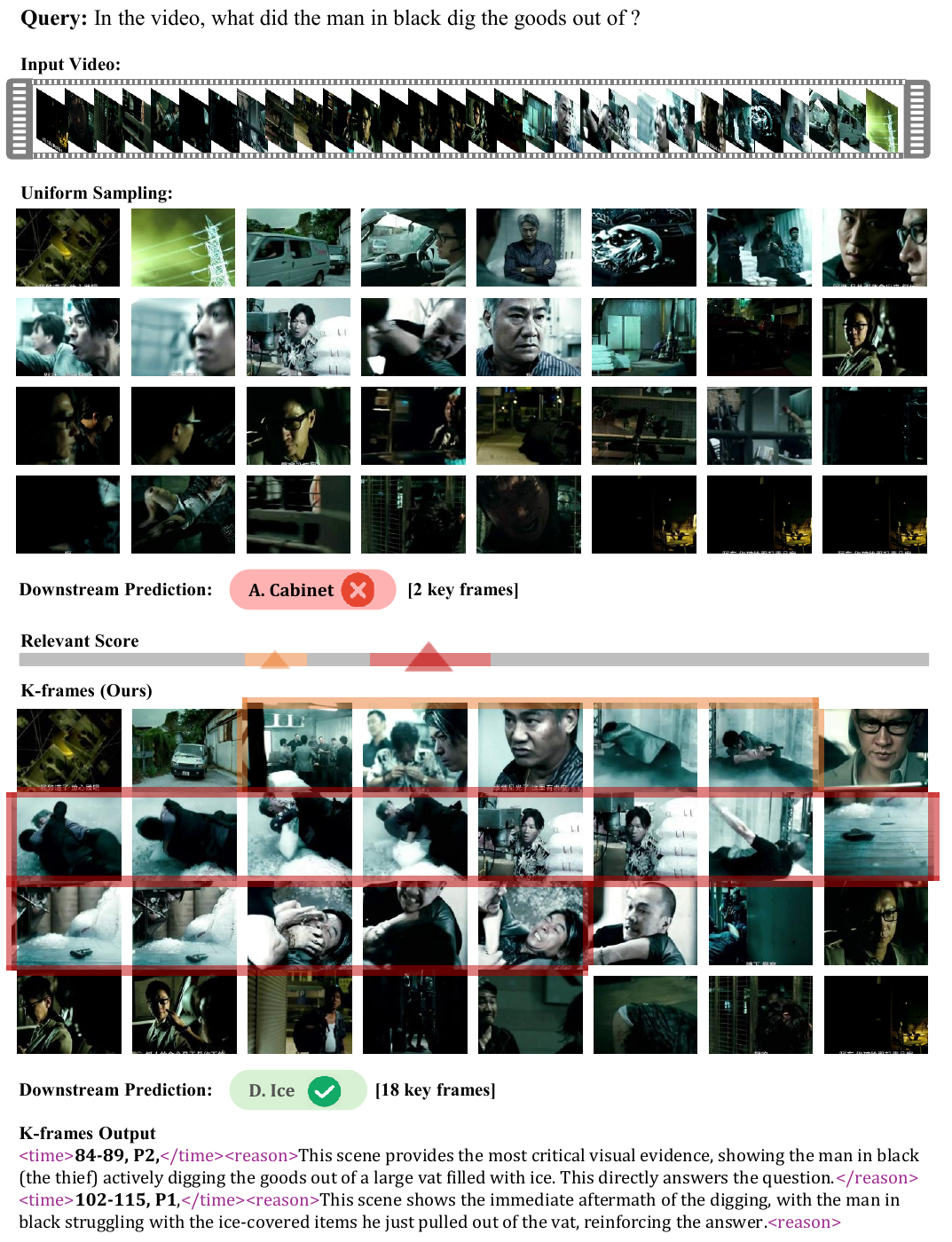} 
\caption{Qualitative comparison between uniform sampling and our K-frames method with the number of frames set to $k=32$.}
\label{fig:cmp_32_1}
\end{figure*}

\begin{figure*}[!t]
\centering
\includegraphics[width=1\textwidth]{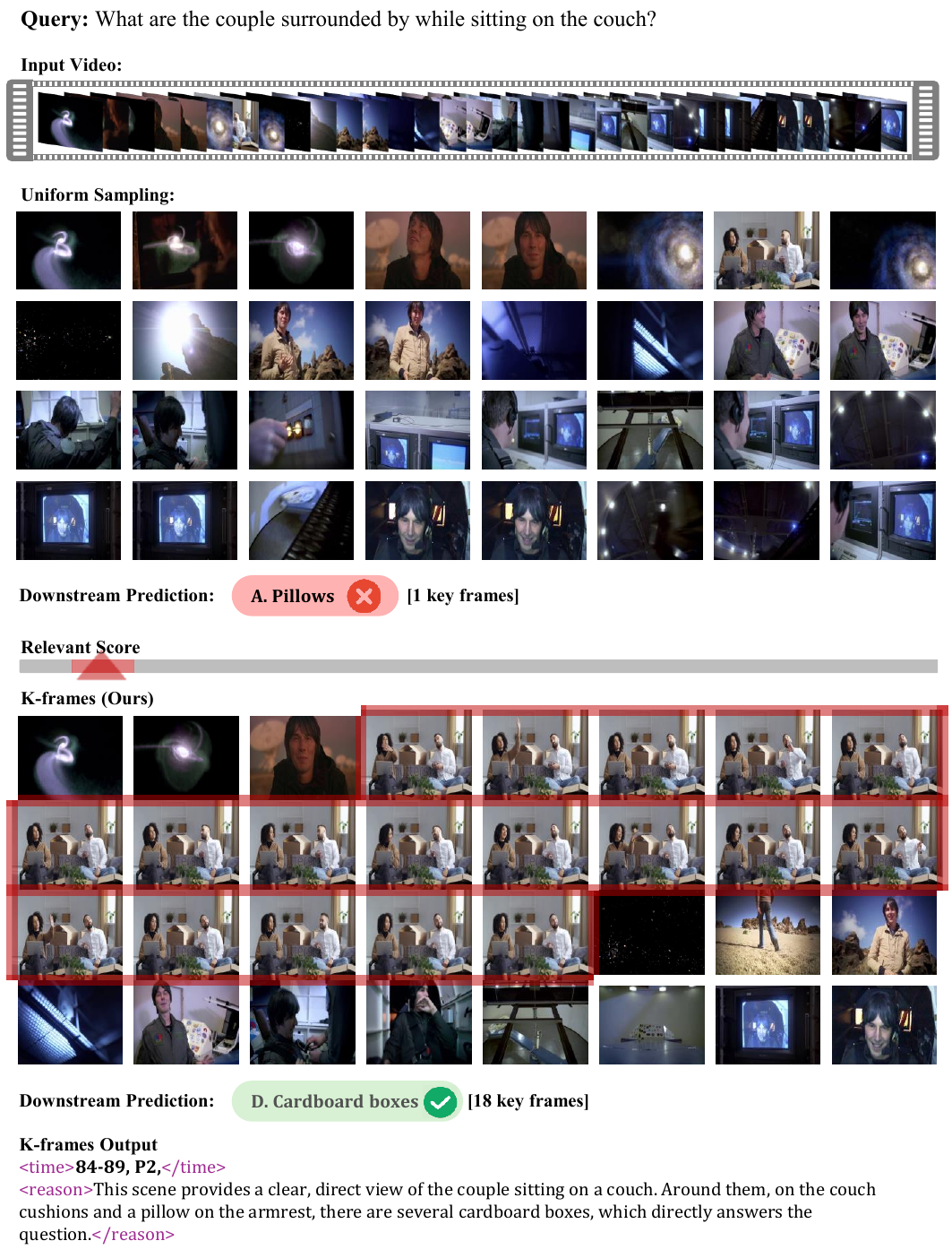} 
\caption{Qualitative comparison between uniform sampling and our K-frames method with the number of frames set to $k=32$.}
\label{fig:cmp_32_2}
\end{figure*}

To further illustrate the robustness and interpretability of our method across different number of frame set, we provide additional qualitative comparisons in this subsection. As shown in Figure~\ref{fig:cmp_8_appendix} and~\ref{fig:cmp_8_appendix_2}, when operating with the number of frame set $k=8$,  K-frames concentrates its selection entirely within the highest-scoring scene clips to capture the most critical visual evidence. When the frame set is increased to $k=32$, K-frames showcases its flexible, multi-scale selection capability. As seen in Figure~\ref{fig:cmp_32_1} and~\ref{fig:cmp_32_2}, our model continues to densely sample the most relevant clips, such as the man digging goods from ice or the couple surrounded by cardboard boxes. It also dynamically allocates a portion of its larger budget to select frames from other, less critical scene clips. This strategy provides a richer and more comprehensive visual context to the LLM while still prioritizing the most query-relevant information, further demonstrating the adaptability of our clip2frame paradigm.

\end{document}

%% file: math_commands.tex
\usepackage{amsmath,amsfonts,bm}

\def\eqref#1{equation~\ref{#1}}

\def\1{\bm{1}}

\def\vb{{\bm{b}}}
\def\vc{{\bm{c}}}

\def\vp{{\bm{p}}}

\def\vs{{\bm{s}}}

\DeclareMathAlphabet{\mathsfit}{\encodingdefault}{\sfdefault}{m}{sl}
\SetMathAlphabet{\mathsfit}{bold}{\encodingdefault}{\sfdefault}{bx}{n}

